\documentclass{article}

\usepackage{amssymb}
\usepackage{amsmath}
\usepackage{hyperref}
\usepackage{natbib}
\usepackage{dsfont}
\usepackage{booktabs}
\usepackage{pifont}
\usepackage{xcolor}

\usepackage{placeins}

\usepackage{mathtools}
\usepackage[inline,shortlabels]{enumitem}
\DeclarePairedDelimiter{\aset}{\{}{\}}

\newif\ifdrafting 
\draftingfalse

\ifdrafting
\newcommand{\GD}[1]{{\textbf{\color{purple}Gideon: #1}}}
\newcommand{\HY}[1]{{\textbf{\color{blue}Hugo: #1}}}
\newcommand{\MH}[1]{{\textbf{\color{red}Matthias: #1}}}
\else
\newcommand{\GD}[1]{}
\newcommand{\HY}[1]{}
\newcommand{\MH}[1]{}
\fi

\newcommand{\bx}{\boldsymbol{x}}
\newcommand{\bz}{\boldsymbol{z}}
\newcommand{\bp}{\boldsymbol{p}}
\newcommand{\bq}{\boldsymbol{q}}
\newcommand{\bs}{\textbf{s}}
\newcommand{\bd}{\textbf{d}}
\newcommand{\lcl}{\mathcal L^{\textsc{CL}}}
\newcommand{\scl}{\mathcal L^{\textsc{SCL}}}
\newcommand{\lagg}{\mathcal{L}^{\textsc{NA}}}
\newcommand{\ldisc}{\mathcal{L}^{\textsc{ND}}}
\newcommand{\ltotal}{\mathcal{L}^{\textsc{NCL}}}
\newcommand{\NCLw}{\textsc{NCL}(n_w)}
\newcommand{\NCLY}{\textsc{NCL}(n_Y)}
\newcommand{\NCLwY}{\textsc{NCL}(n_{w \cap Y})}

\usepackage{microtype}
\usepackage{graphicx}
\usepackage{subfigure}
\usepackage{booktabs} \usepackage{amsmath,amsthm,amsfonts}
\usepackage{atbegshi}

\usepackage{hyperref}

\usepackage[accepted]{icml2021}

\icmltitlerunning{Neighborhood Contrastive Learning Applied to Online Patient Monitoring}

\begin{document}

\twocolumn[

\icmltitle{Neighborhood Contrastive Learning Applied to Online Patient Monitoring}

\icmlsetsymbol{equal}{*}

\begin{icmlauthorlist}
\icmlauthor{Hugo Yèche}{equal,eth}
\icmlauthor{Gideon Dresdner}{equal,eth}
\icmlauthor{Francesco Locatello}{amazon}
\icmlauthor{Matthias Hüser}{eth}
\icmlauthor{Gunnar Rätsch}{eth}
\end{icmlauthorlist}

\icmlaffiliation{eth}{Department of Computer Science, ETH Zürich, Switzerland}

\icmlaffiliation{amazon}{Amazon (most work was done when Francesco was at ETH Zurich and MPI-IS)}

\icmlcorrespondingauthor{Hugo Yèche}{hyeche@ethz.ch}
\icmlcorrespondingauthor{Gideon Dresdner}{dgideon@ethz.ch}

\icmlkeywords{ Representation learning, Healthcare, Unsupervised Learning}
\vskip 0.3in
]

\printAffiliationsAndNotice{\icmlEqualContribution}

\begin{abstract}
Intensive care units (ICU) are increasingly looking towards machine learning for methods to provide online monitoring of
critically ill patients. In machine learning, online monitoring is often formulated as a supervised learning problem. Recently,
contrastive learning approaches have demonstrated promising improvements over competitive supervised benchmarks. These
methods rely on well-understood data augmentation techniques developed for image data which do not apply to online monitoring. In 
this work, we overcome this limitation by supplementing time-series data augmentation techniques with a novel contrastive
learning objective which we call neighborhood contrastive learning (NCL). Our objective explicitly groups together 
contiguous time segments from each patient while maintaining state-specific information. Our experiments demonstrate
a marked improvement over existing work applying contrastive methods to medical time-series.
\end{abstract}
 \vspace*{-3ex}
\section{Introduction}

Recent advances in contrastive learning have shown that unsupervised learning can improve upon 
competitive computer vision benchmarks \citep{misra2020self,chen2020simple,he2020momentum,tian2020makes,caron2020unsupervised}. Such methods rely on data augmentations to construct semantic-preserving views of samples. By aggregating them using a Noise Contrastive Estimation  objective \citep{gutmann2010noise}, contrastive learning aims to learn view-invariant representations. In subsequent work, \citet{khosla2020supervised} showed that a supervised extension of this objective would further extend its benefits over end-to-end training. Building on these successes, researchers applied this methodology to medical time-series data \citep{cheng2020subject,kiyasseh2020clocs,mohsenvand2020contrastive}.

Supervised learning approaches have enjoyed successes in online monitoring of organ failure and other life-threatening events \citep{hyland2020early,tomavsev2019clinically,schwab2020real,horn2019, li2020time}. On the other hand, unsupervised representation learning has not been widely applied in this setting. To our knowledge, \citet{lyu2018improving} is the only work that makes an attempt. We conjecture that this is due to the additional challenges of working in this setting: difficult-to-interpret datatypes and heterogeneous distributions of samples.

Time-series are often less humanly understandable than images. Finding semantically preserving augmentations --- crucial to recent advances in contrastive learning --- is challenging \citep{um2017data, fawaz2018data}. In addition, biosignal data suffers from a particular domain heterogeneity problem due to multiple samples originating from the same patient. While samples from a single patient will have many commonalities, they also will exhibit changes which become important when trying to perform learning-based prediction \citep{morioka2015learning,farshchian2018adversarial,ozdenizci2020learning}. In the online monitoring setting, these shifts are amplified by large overlaps in history between time segments from a single patient stay.

Preliminary works \citep{cheng2020subject,kiyasseh2020clocs} propose sampling approaches to overcome this between-patient heterogeneity in a contrastive learning setup. However, their methods only cover the edge cases of complete dependence or non-dependence between labels and source subjects, while ignoring the factor of time. Such assumptions show limitations in online monitoring where a patient state evolves continuously. In this work, we propose a contrastive learning framework addressing the complexity of online monitoring tasks\footnote{\url{https://github.com/ratschlab/ncl}}. Our contributions can be summarized as follows:\vspace*{-2ex}
\begin{enumerate}[itemsep=-1ex]
\item  
We propose NCL, a new contrastive learning objective that can explicitly induce a prior structure to representations in a modular, user-defined manner, allowing to address the heterogeneity problem in online monitoring data.
\item
We show that, when used unsupervised, our approach shows competitive results for several online monitoring benchmarks on open-source medical datasets. In addition, it outperforms supervised learning in the limited-labeled-data setting.
\item
Moreover, when supervised, we demonstrate that it outperforms end-to-end counterparts and previous supervised extensions to contrastive learning. We also show significant improvement over them in a transfer learning setting.

\end{enumerate}

\section{Related Work}

\paragraph{Contrastive learning for biological signals.}There is a recent burst of work applying unsupervised contrastive learning to 
biosignal data such as electroencephalogram (EEG) and electrocardiogram (ECG)  
\citep{banville2020uncovering,kiyasseh2020clocs,cheng2020subject,mohsenvand2020contrastive}. The problems posed by 
such biosignal data have important commonalities and differences with EHR data. Unlike EHR data, biosignal data has 
a much higher time resolution and known spatial dependencies between channels. 

Like EHR data, biosignal data exhibits between-patient heterogeneity. We show in Section \ref{sec:ps_model} that for online monitoring data this problem is further aggravated as the within-patient variance of 
time points is smaller than the between-patient variance.
\citet{cheng2020subject} tackle this problem by limiting their negative sampling procedure to samples only within 
the same patient, whereas \citet{kiyasseh2020clocs} do the opposite --- by exclusively sampling negatives across patients. 
Both \citet{banville2020uncovering} and \citet{franceschi2019unsupervised} approach this problem by enforcing 
temporal smoothness between contiguous samples, similar in spirit to \citet{mikolov2013efficient}.

\citet{mohsenvand2020contrastive} do not pay special attention to between-patient heterogeneity. Instead, they achieve competitive 
results by applying recently improved neural network architectures and augmentation techniques from
\citet{chen2020simple}. \citet{mohsenvand2020contrastive} provide the impetus for our work. We expand the
methodology of \citet{chen2020simple} and \citet{he2020momentum} to incorporate a contrastive 
learning objective capable of dealing with the patient-induced heterogeneity problem.

\paragraph{Unsupervised patient state representation for EHR Data.} There is a growing body of work on unsupervised 
representation learning that learns patient-wise representations 
\citep{miotto2016deep,darabi2019unsupervised,landi2020deep}. While EHR data is often in the form of a time-series, this 
body of work focuses on learning a single, time-invariant representation per patient, which cannot support online
monitoring tasks.

In contrast, there is little prior work on time-dependent patient representations. To the best of our knowledge, the Seq2Seq autoencoder-based work of \citet{lyu2018improving} is the only one. Their work aims to further develop patient state representation with the goal of improving fine-tuning performance in the limited labeled data setting. Other works also explore patient state-representation learning but with specific focuses such as reinforcement learning \cite{killian2020empirical} or multi-task learning \cite{mcdermott2020comprehensive}.  
\section{Neighborhood Contrastive Learning (NCL)}\label{sec:NCL}

\subsection{Preliminaries} \label{sec:definitions}

\paragraph{Data definition.}
We are given a set of patients where each patient may have multiple ICU stays. Taking the union across all patients gives a total of $S$ patient stays. Each patient stay itself is composed of a vector of static demographic features $\bd^p$, known at admission time, and a multivariate time-series $\bs^p$ describing the stay.

Online monitoring aims to make a prediction given the history up to time $t$. Because the length-of-stay can vary considerably across patients, we define a maximum history window $t_h$ and slide that window across the patient's time series. Each window is denoted as $\bs_t^p$ = $[s_{t-t_h}^p,\ldots,s_{t}^p]$. Our goal is to make a prediction for each $\bx_t^p \triangleq (\bd^p,\bs_t^p)$.

Each patient stay has total history length of $t^p$. Taking the union of all windowed time-segments $\mathcal{S}^p = \{(\bd^p,\bs^p_t) \mid t \leq t^p \}$ gives the final dataset definition $\mathcal{D} = \bigcup^S_{p=0} \mathcal{S}^p$.

\paragraph{Pipeline specification.}
\GD{we have already defined $\bx$, now we just need to stay that $\bp$ is the output. }
\GD{Not definition, that will be elsewhere. Here we merely specify the inputs/outputs.}

Let $B=\aset{\bx_1,\ldots,\bx_N}$ denote a minibatch of $N$ examples.
To each example we associate two views $\tilde\bx_i$ and $\tilde\bx_{v(i)}$ which are constructed using data augmentation. 
This gives a total of $2N$ views in the mini-batch $V=\aset{\tilde{\bx}_1,\ldots, \tilde{\bx}_{2N}}$.

To each view, we apply an encoder and a momentum encoder, denoted $f_e$ and $f_m$ respectively, to obtain 
representations $Z^e = \aset{\bz_1^e,\ldots,\bz_{2N}^e}$ and $Z^m = \aset{\bz_1^m,\ldots,\bz_{2N}^m}$. All representations
are normalized by dividing by their Euclidean norms (projecting onto the unit sphere) 
\citep{he2020momentum,tian2020makes,wang2020understanding}.

During training, these representations are further projected using two distinct projectors, $h_e(\cdot)$ for $Z^e$ and $h_m(\cdot)$ for $Z^m$ resulting in two sets of projected representations, $P = \{\boldsymbol{p}_1,\ldots \bp_{2N}\}$ and $P^m = \{\boldsymbol{p}^m_1,\ldots,\bp_{2N}^m\}$. These projections are also normalized by their Euclidean norms.

$P^m$ is used to update a queue of negative samples $Q~=~\aset{\bq_1,\ldots,\bq_M}$. Specifically, at each training step, the oldest $2N$ elements of $Q$ are replaced by $P^m$ in sliding manner. Particularly, for $k < 2N$, we have $q_k = p^m_k$. This allows for a large number of negative samples since we can choose $M\gg2N$.

\subsection{Regular Contrastive Loss}
Consider the contrastive objective from \citet{chen2020improved}.
\begin{equation} \label{eq:CL}
\lcl=-\sum_{i=1}^{2N} \log \frac{\exp\left(\boldsymbol{p}_{i} \cdot \boldsymbol{p}^m_{v(i)} / \tau\right)}{\sum_{k \neq i}^M \exp{\left(\boldsymbol{p}_{i} \cdot \boldsymbol{q}_{k} / \tau\right)}}
\end{equation}
where $\tau>0$ is the temperature scaling parameter.

This objective does not take labels into account. As a result, it is possible that a sample $\bx_k$ with the same label as $\bx_i$ is selected as a negative sample which repels their corresponding representations. \citet{khosla2020supervised} proposes a supervised objective to remedy this issue. They define a loss which gives an attractive force to representations generated from examples with the same label. Their so-called Supervised Contrastive Loss (SCL) is defined as 

\begin{equation} \label{eq:SCL}
 \scl= \sum_{i=1}^{2N} \frac{-1}{|Y(i)|} \sum_{l\in Y(i)} \log \frac{\exp \left(\boldsymbol{p}_{i} \cdot \boldsymbol{p}^m_{l}/\tau \right)}{\sum_{k \neq i}^M \exp \left(\boldsymbol{p}_{i} \cdot \boldsymbol{q}_{k} / \tau\right)}
\end{equation}

where $Y(i) = \aset{j \mid y_j = y_i}$ indexes samples with the same label as $\bx_i$.

There are two important observations to make. First, $Y(i)$ is defined arbitrarily in terms of labels, therefore the learned representation is specific to a given downstream task.

Second, SCL tends towards a degenerate solution where representations of examples with the same label collapse to become the same vector. This might lead to poor generalization to other tasks, thus reducing the performance of SCL in transfer learning. In the following section we go through the steps to generalize this term and derive the corresponding training objective.

\subsection{Neighborhood-aware Loss}
\HY{Changed name of $\lagg$ and $\ldisc$}

We say that two samples share the same neighborhood if they share some predefined attributes. For example, two words pronounced by the same speaker in speech recognition or two states of a patient that are within $w$ hours of each other in an EHR.

We represent this as a collection of binary functions of the form $n(\bx_i,\bx_j)$ which returns one when $\bx_i$ and $\bx_j$ are in the same neighborhood and zero otherwise. These functions provide a modular way of defining neighborhoods. The $i$-th neighborhood $N(i)=\aset{k\ne i\mid n(\bx_i,\bx_k)=1}$ is simply the set of samples for which $n(x_i,\cdot)$ returns one. Note that $n(\bx_i,\bx_j) = \mathds{1}[y_i = y_j]$ gives precisely $N(i) = Y(i)$ as defined in \citet{khosla2020supervised}.

This naturally leads to our generalization of $\scl$ which we call the Neighbors Alignment (NA) objective. This objective encourages representations coming from the same neighborhood to align.

\begin{equation} 
\lagg
=\sum_{i=1}^{2N} \frac{-1}{|N(i)|} \sum_{l \in N(i)} \log \frac{\exp\left(\boldsymbol{p}_{i} \cdot \boldsymbol{p}^m_{l} / \tau\right)}{\sum_{k \neq i}^M \exp \left(\boldsymbol{p}_{i} \cdot \boldsymbol{q}_{k} / \tau\right)}
\end{equation}
\normalsize

By itself, $\lagg$ suffers from the same drawbacks as $\scl$. Specifically, it will eventually result in a trivial solution in which neighborhoods collapse to a single point. To remedy this problem, we propose the Neighbor Discriminative (ND) objective: 
\begin{equation}
        \ldisc=-
        \sum_{i=1}^{2N}\log \frac{\exp \left(\boldsymbol{p}_{i} \cdot \boldsymbol{p}^m_{v(i)} / \tau\right)}{\sum_{k \in N(i)} \exp \left(\boldsymbol{p}_{i} \cdot \boldsymbol{q}_{k} / \tau\right)}
\end{equation}
\normalsize
This objective more closely resembles the original contrastive learning objective described in \citet{chen2020simple}. It encourages representations from views of the same anchor to be similar to one another. Though, negatives used for normalization are neighbors to the anchors. This objective allows preserving a needed diversity within each neighborhood, as neighbors do not necessarily share the same downstream task label.

Our final objective is a weighted average between these two objectives called \emph{Neighborhood Contrastive Learning}. We introduce an explicit trade-off parameter $\alpha\in [0,1]$ to smoothly interpolate between intra- and inter-neighborhood properties of the representations:
\begin{equation} \label{eq:NCL}
\ltotal = \alpha \lagg + (1-\alpha) \ldisc
\end{equation}

\begin{figure}[H]
\vspace*{-3ex}
\begin{center}
\centerline{\includegraphics[width=\columnwidth]{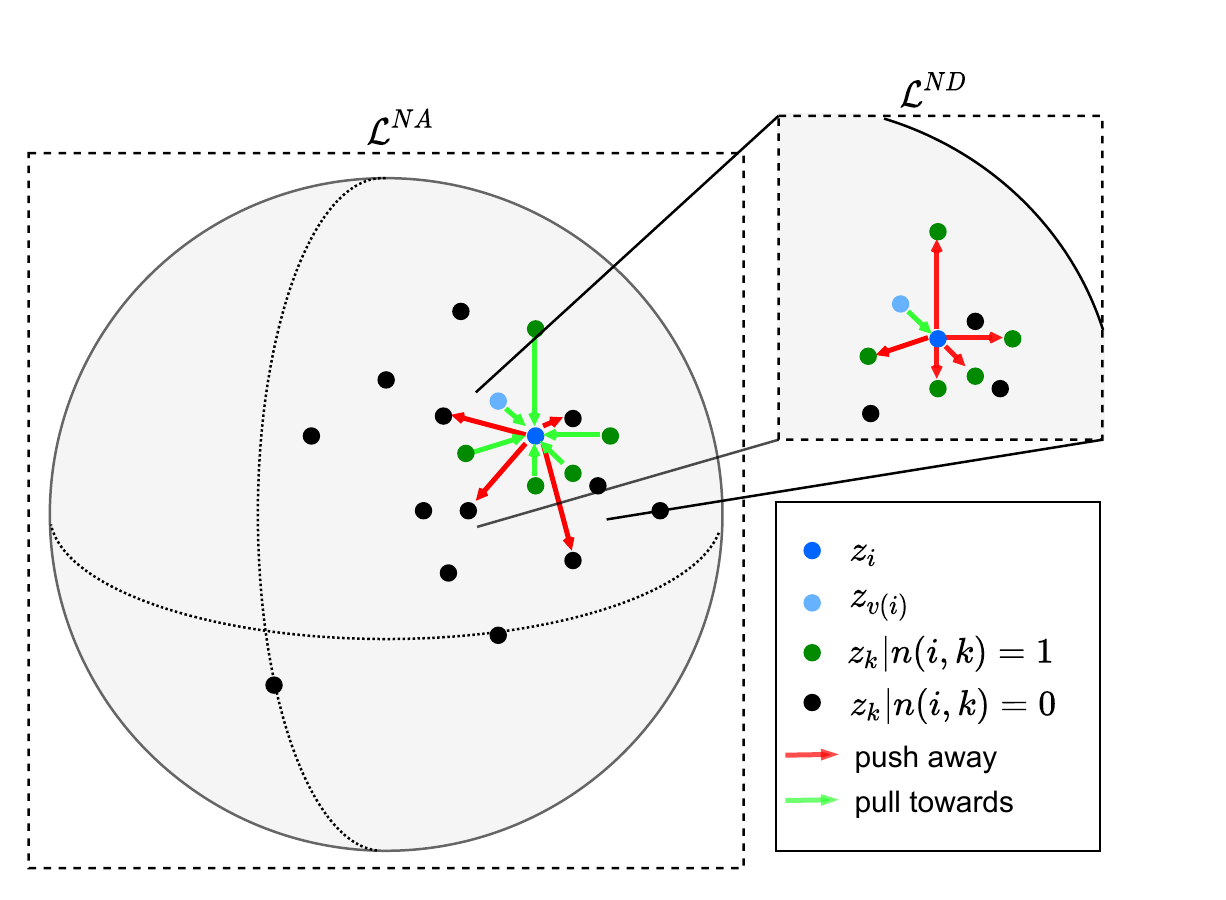}}
\caption{Illustration of our objective $\ltotal$. On the left, $\lagg$ induces embeddings from the same neighborhood (green) as the anchor $\bz_i$ (dark blue) to be closer than samples external to it (black). In contrast, on the right, $\ldisc$ preserves a hierarchy between neighbors (green) and the other view of the anchor's original example $z_{v(i)}$ (light blue).}
\label{fig:NCL}
\end{center}
\vskip -0.2in
\end{figure}

\begin{figure*}[t]
\vskip 0.2in
\begin{center}
\centerline{\includegraphics[width=\textwidth]{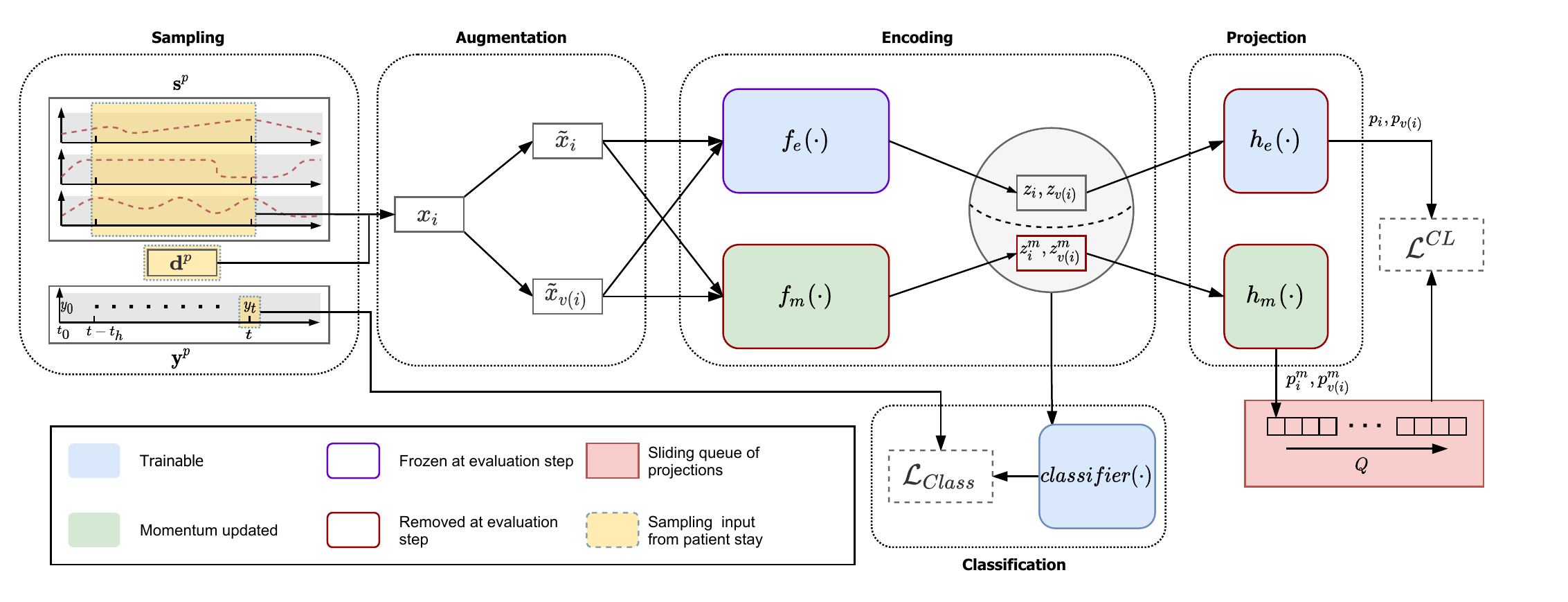}}
\caption{ Schema of the contrastive pipeline initially proposed by \citep{chen2020improved}. From a patient stay $p$, we sample $x_i = (\bs^p_t, \bd^p)$ corresponding to the patient state at time $t$. We augment it twice and pass both views, $\tilde{x_i}$ and  $\tilde{x_{v(i)}}$, through an encoder $f_e$ and a momentum encoder $f_m$. At training time, the representations are further projected with $h_e$ and $h_m$. From these projections and the  sliding momentum queue $Q$, we compute the contrastive objective $\lcl$. At evaluation time, we freeze $f_e$ and train a classifier on top of the learned representation. }
\label{fig:ssl-pipe}
\end{center}
\vskip -0.2in
\end{figure*}

\section{Application to Online Monitoring} \label{sec:ps_model}

\HY{TODO : Do a pass on this section once we finalized definitions}
\subsection{Motivations} \label{om-motivation} 

Introduced as in Section \ref{sec:NCL}, our framework, as a combination of a neighborhood function and an objective, is quite general.  While many applications exist, we believe online monitoring data shows two specific attributes making it the better candidate to motivate its use.

Using the definition from Section \ref{sec:definitions}, we note that each dataset $\mathcal{D}$ is composed of subsets of samples $\mathcal{S}^p$, all originating from a single patient stay. Moreover, elements from these subsets not only share their origin but are all time-dependent on one another. Therefore, there exists a prior structure to the data unknowingly correlated to its labeling. When using a patient-dependant neighborhood, $\lagg$ aims to preserve the patient-specific features common to all neighbors, while $\ldisc$ aims to find discriminative features across neighbors. 

The previous observations motivates the use of our method.
However, other data types collected from human subjects, including biosignals, exhibit the same attributes. Online monitoring distinguishes itself by the existence of overlaps between examples from contiguous states of a patient. These shared history segments not only increase prior dependencies among samples, but they also limit the possible use of temporal data augmentations to mitigate the problem. Indeed, any augmentation creating identical views from two separate examples does not preserve their singularity.

To summarize, as other data originating from human subjects, online monitoring data shows strong distribution shifts across patients unknowingly correlated to labeling. Yet, relying on augmentation strategies to tackle the problem is limited by their need to be semantic-preserving. That motivates focusing on other components to remedy this issue as proposed in our work.

\subsection{Design of neighborhood functions for online monitoring}
Based on our previous observations of the online monitoring of patients state, we now detail our two proposed neighborhood functions to cope with distribution heterogeneity: \begin{enumerate*}[(1)]
    \item  time-preserving 
    \item  label-preserving neighborhood 
\end{enumerate*}.

The first preserves the time dependency of the representations of the time-series segments. We chose to consider as neighbors samples from a patient that are close in time motivated by \cite{banville2020uncovering} and \cite{franceschi2019unsupervised} works.  We define a neighborhood function $n_w(\bx_i,\bx_j)$ with window size $w$. To samples $\bx_i$ and $\bx_j$, we can associate segments from patient stays, $\bs^{p_i}_{t_i}$ and $\bs^{p_j}_{t_j}$. Then $n_w(\bx_i,\bx_j)$ is an indicator function which takes the value one when $p_i=p_j$ \emph{and} $|t_i-t_j|<w$ and zero otherwise. The associated neighborhood $N_w(i)$ is defined as in Section \ref{sec:NCL}. We refer to this approach as $\NCLw$.

The label-preserving neighborhood function is $n_Y(\bx_i,\bx_j)$ which takes the value of one when the two samples have the same downstream label and zero otherwise. This resembles \citet{khosla2020supervised} but, when inserted into our proposed contrastive learning objective (Eq.\ \ref{eq:NCL}) still provides a balance between label-specific and general features. We refer to the resulting learning objective as $\NCLY$.

\subsection{Contrastive Framework} \label{contrastive-framework}

We adopt the pipeline used by \citet{cheng2020subject,chen2020improved,chen2020big}, depicted in 
Figure \ref{fig:ssl-pipe}. Positive pairs for medical time-series are constructed using extensions of existing data augmentation techniques
described in \citet{cheng2020subject,kiyasseh2020clocs,mohsenvand2020contrastive}. Negative sampling is implemented 
via a momentum encoder \citep{he2020momentum}. Finally, standard projection layers are incorporated as the final 
layers of the architecture \citep{chen2020simple}.
As in \citet{he2020momentum}, we define an in-place momentum update. At each training step $t$, $f_m = (1-\rho)f_e + \rho f_m$ and similarly $h_m = (1-\rho)h_e + \rho h_m$ for hyperparameter $\rho\in (0,1)$.

\paragraph{Patient-state encoder.}
Building on previous work in contrastive learning on time-series data, we use Temporal Convolutional Networks (TCN)
\citep{bai1803empirical}. We slightly modify the original TCN architecture using layer normalization \citep{ba2016layer}. 
We also add a dense layer after the TCN block to merge the static features from each patient into the final state
representation. We refer to the Appendix~\ref{appendix_data} Figure \ref{fig:architeture} for a diagram of the architecture.

\paragraph{Data augmentations for online monitoring.} \label{section_da}
We use channel dropout \citep{cheng2020subject} and Gaussian noise \citep{kiyasseh2020clocs, mohsenvand2020contrastive} to create positive pairs. In channel dropout, we randomly mask out a subset of variables while keeping the full time-series intact. Gaussian noise is simply the addition of independent Gaussian noise to each variable. Concerning the demographic vector in each sample, we only use random dropout. We did not use other channel augmentations from these approaches due to their specificity to EEG/ECG.

Using temporal augmentations from previous work is not straight forward. As explained in Section \ref{om-motivation}, contrary to EEG/ECG, large history overlaps exist between samples from adjacent states of a patient stay. Therefore randomly shifting or cropping the time-series component of these samples can lead to identical augmented views. Inspired by \citet{cheng2020subject}'s work, we introduce two augmentations that never alter the last step of a time-series, preserving its singularity in the online monitoring context. First, history cutout, which masks out a short window from the time-series but excludes the last step from the possible candidates. Second, history crop, which crops a time-series along the temporal axis. However, we only crop from the past. More details about the data augmentations can be found in Appendix~\ref{appendix_data}. 
\section{Experimental Setup}

\subsection{Datasets}
\HY{I did a pass on this subsection, what do you think @Gideon ?}
To benchmark the proposed method and allow further work to compare to our results, we selected two well-known EHR datasets that are openly available and for which online monitoring tasks exist.

\paragraph{MIMIC-III Benchmark.}
The MIMIC-III dataset \citep{johnson2016mimic} is the most commonly used dataset for tasks related to EHR data. However, as noted by \citet{bellamy2020evaluating}, many previous approaches using it, applied their custom-built pre-processing pipeline and selection of variables, and thus making a comparison across methods impossible. To allow further comparison to our work, we used the pre-processed version of the dataset by \citet{harutyunyan2019multitask}, referred to as ``MIMIC-III Benchmark''. Among the four tasks they define, we used the two with hourly labeling for each patient stay, \textit{Decompensation} and \textit{Length-of-stay} predictions.

\textit{Decompensation} is a binary classification task which aims to predict whether a patient at time $t$, is going to pass away in the upcoming 24h. This task is highly unbalanced. Thus, as in  \citet{harutyunyan2019multitask}, we evaluate it with AUROC and AUPRC metrics.

For the \textit{Length-of-stay} task, we aim to predict the remaining time the patient will stay in the ICU. This task is a regression problem. However, due to the heavy-tailed distribution of labels,  \citet{harutyunyan2019multitask} frame it as a 10-way classification where each class is a binned duration of time. To evaluate the method, like them, we use linear weighted Cohen's Kappa. The score is between -1 and 1, 0 corresponding to random predictions.

The benchmark dataset extracted from MIMIC-III by \citet{harutyunyan2019multitask} contains more than 50,000 stays across 38,000 distinct patients for which 17 measurements are provided. More details about the dataset can be found in Appendix \ref{appendix_ds}.

\paragraph{Physionet   2019.}
This dataset originates from a challenge on the early detection of sepsis from clinical data \citep{reyna2019early}. It contains more than 40,000 patients from two hospitals, for which a total of 40 variables are available. Because the original test set was not available, we used the splits of \citep{horn2019} on the openly available sub-set of patients.

The task is to hourly predict sepsis onset occurring within the next 6h to 12h. To evaluate performances, we also used the novel ``Utility'' metric introduced by the authors of the challenge  \citep{reyna2019early}. Compared to AUROC and AUPRC, this metric is more clinically relevant as it penalizes differently false predictions depending on their relative temporal distance to a sepsis event. A score of zero corresponds to a classifier predicting no sepsis event. For a perfect classifier, the maximum score is one.

\begin{table}[tbh]
\caption{Other contrastive methods as instances of our framework.}
\label{tab:baseline}
\vskip 0.1in
\begin{center}
\begin{small}
\begin{tabular}{@{}lccc@{}}
\toprule
Method & $\alpha$ & $w$ & $n(\cdot,\cdot)$ \\
\midrule
CL & $1.0$& $0$ & $n_w$  \\
SACL \citep{cheng2020subject} & $0.0$& $ +\infty$ & $n_w$   \\
CLOCS \citep{kiyasseh2020clocs} & $1.0$& $ +\infty$ & $n_w$   \\
SCL \citep{khosla2020supervised} & $1.0$& N.A & $n_Y$ \\
\bottomrule
\end{tabular}
\end{small}
\end{center}
\vskip -0.1in
\end{table}

\subsection{Baselines}
To achieve a fair comparison, we compare all approaches using the same encoder detailed in Section \ref{set_up}. Like others, we compare our method to the so-called ``End-to-end'' baseline. It consists of training the identical architecture in a supervised manner using downstream task labels.

We also compare our work to existing contrastive approaches reported in Table \ref{tab:baseline}. All of them fall under different settings of our general framework. Regular CL, as a direct adaptation of \citet{chen2020improved,chen2020big} work, can be reproduced using a window size $w =0$ and only $\lagg$. Specific methods for medical time-series, SACL  ``Subject-specific'' sampling method \citep{cheng2020subject} and CLOCS \citep{kiyasseh2020clocs}, are equivalent to using $w = +\infty$, with respectively only $\ldisc$ or $\lagg$. Finally, SCL \citep{khosla2020supervised} is similar to CLOCS, except its neighborhood function is defined based on the downstream task labels, making the method supervised.

We also compare the unsupervised version of our framework, using $n_w$, to Seq2Seq auto-encoders. First, an auto-encoder (Seq2Seq-AE), trained to minimize the mean square error (MSE) over the input time segment and static vector. Then, a forecasting one (Seq2Seq-AE-forecast), which defines the reconstruction loss over the consecutive time segment of the same length.

\subsection{Implementation} \label{set_up}

\paragraph{Pre-processing.}
First, for each dataset, we re-sampled each stay to hourly resolution and filled missing values with forward-filling imputation. We then applied standard scaling to each non-categorical variable based on the training set statistics. We also one-hot encoded the remaining categorical variables. Finally, we zero-imputed (corresponding to the mean of the training set after scaling) the remaining missing values after the forward-filling imputation. We used a maximum history length $t_h$ of 48 hours. We pre-padded shorter time-series. This pre-processing leads to input of size $48 \times 42$ for MIMIC-III Benchmark tasks and $48 \times 40$ for the Physionet 2019 one.

\paragraph{Architecture.}
From architecture searches (Appendix \ref{hp_appendix}) for the end-to-end baseline, we used the common encoder depicted in Figure \ref{fig:architeture}, for both data sets. We use a convolution kernel of size 2 and 64 filters. To obtain a receptive field of at least 48 h, we stack five dilated causal convolution blocks. The final embedding dimension after incorporating the static feature is 64.

\paragraph{Unsupervised training parameters.}
We trained all unsupervised methods for 25k steps with a batch size of 2048. We used an Adam optimizer with a linear warm-up between 1e-5 and 1e-3 for 2.5k steps followed by cosine decay schedule as introduced by \citet{chen2020simple}. We selected the common contrastive parameters from performances on the validation set for $\lcl$ objective. More details can be found in Appendix \ref{hp_appendix}. We used a temperature of 0.1, a queue of size 65536, and an embedding size of 64 for all tasks. We set the momentum to 0.999  for MIMIC-III Benchmark tasks and 0.99 for Physionet 2019. Concerning parameters specific to our method, for $\NCLw$ we chose $\alpha = 0.3$ and $w = 16 $ on MIMIC-III Benchmark and $\alpha = 0.4$ and $w = 12 $ on Physionet 2019. For $\NCLY$, we use $\alpha = 0.9$ for all tasks. These parameters were selected using grid searches reported in Appendix~\ref{hp_appendix}. For auto-encoding methods, we used a decoder with a mirrored architecture to the common encoder. However, we did not normalize the representations to the unit sphere.

\paragraph{Model evaluation parameters.}
We evaluated all representation learning methods on a frozen representation. As discussed in Section \ref{sec:mlp_vs_linear} we used two different classification heads, a linear and a non-linear MLP. We used early stopping on validation set loss and an Adam optimizer. The learning rate was set to 1e-4 for all tasks and classification heads on the MIMIC-III benchmark dataset. For Physionet 2019 we used a learning rate of 1e-4 for linear classification and 5e-5 for the MLP head. For the end-to-end baseline, which trains the encoder and classification head simultaneously, we used a smaller learning rate of 1e-5. 
\section{Results}

\begin{figure}[thb]
\vskip 0.2in
\begin{center}
\centerline{\includegraphics[scale=0.45]{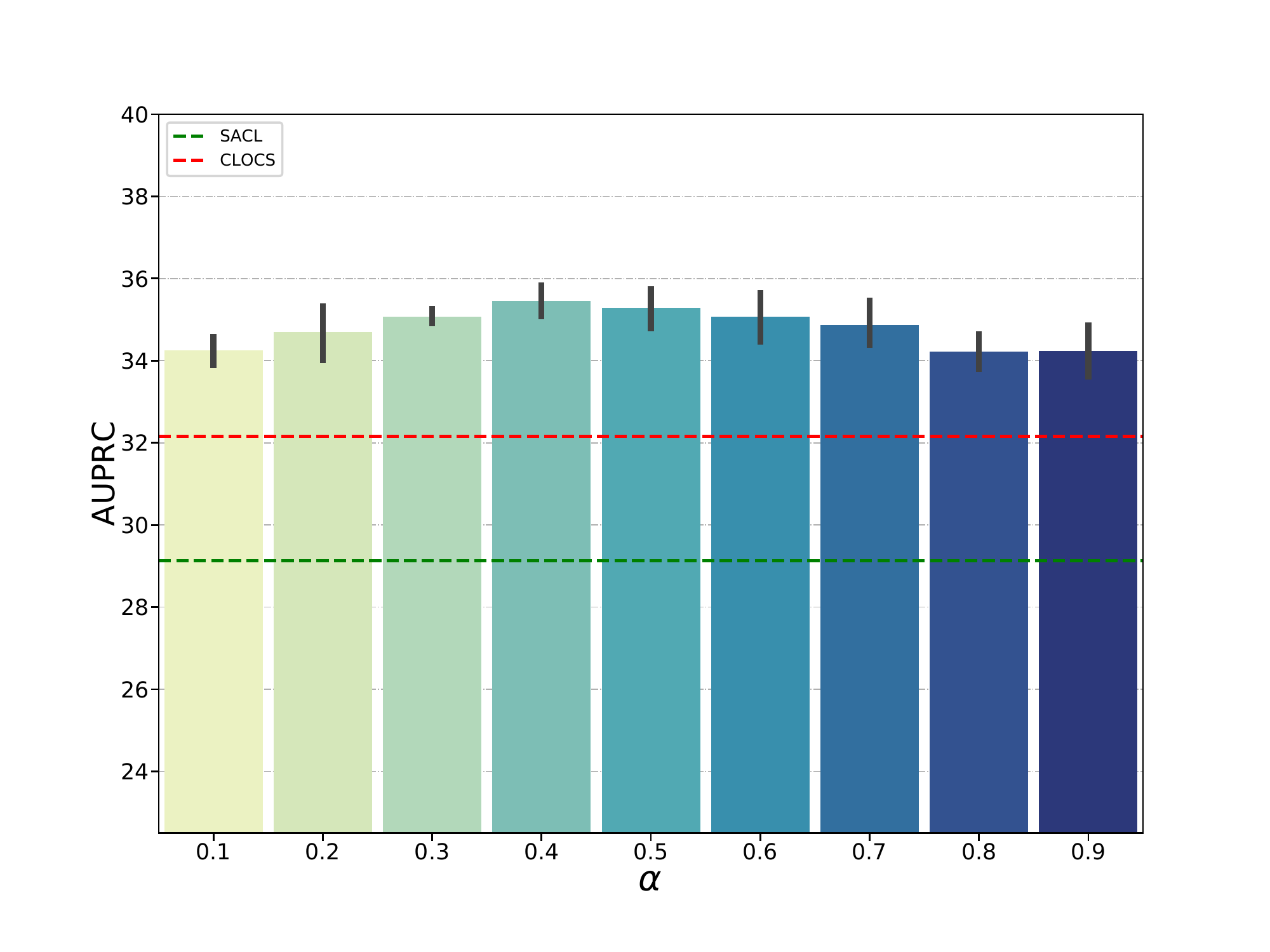}}
\caption{Performance on the \textit{Decompensation} task for various values of $\alpha$ in $\NCLw$. Results are averaged over 5 runs and $w = 16$. We observe a trade-off in performances when varying aggregation as conjectured in Section \ref{sec:NCL}.}
\label{fig:perf_trade-off}
\end{center}
\vskip -0.2in
\end{figure}

\begin{figure}[thb]
\vskip 0.2in

\begin{center}
\centerline{\includegraphics[width=\columnwidth]{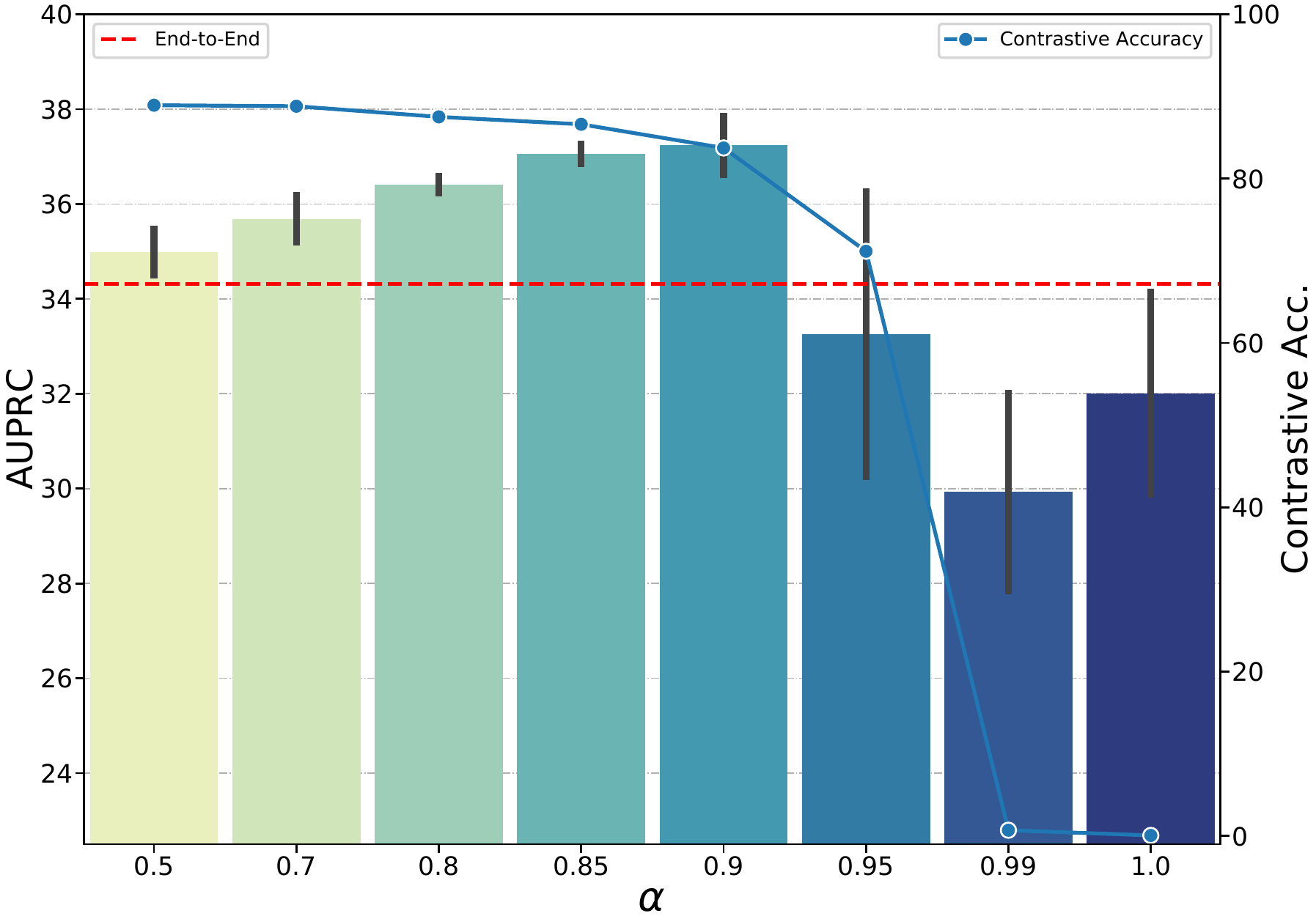}}
\caption{Performance on the \textit{Decompensation} task for various values of $\alpha$ in $\NCLY$. Results are averaged over 5 runs and trained with the \textit{Decompensation} labels. In blue, we show the contrastive accuracy. For $\alpha > 0.95$ it drops because pre-text task becomes too hard explaining SCL's low performances. }
\label{fig:perf_trade-off-ncly}
\end{center}
\vskip -0.2in
\end{figure}

Before discussing results individually, one general thing to note is that end-to-end baselines are competitive with previous work. On the MIMIC-III Benchmark, it performs on par with the proposed methods by \citet{harutyunyan2019multitask}. Similarly, for the same splitting of Physionet 2019, it outperforms all models from \citet{horn2019}.

\begin{table*}[tbh!]
\caption{Results on the MIMIC-III Benchmark dataset. (Top rows) Unsupervised methods; (Bottom rows) Supervised methods. All scores are averaged over 20 runs such that the reported score is of the form $mean \pm std$. In bold are the methods within one standard deviation of best one for each setting. Evaluation metrics were scaled to 100 for readability purposes. (D) and (L) stands for Decompensation and Length-of-Stay indicating which labels were used to train the representation. To get competitive results we had to froze the projector for SCL. }
\vskip 0.1in
\begin{center}
\begin{small}
\begin{tabular}{l|ll|ll||ll}
\toprule
Task & \multicolumn{4}{|c||}{Decompensation} & \multicolumn{2}{c}{Length-of-stay} \\
\midrule

Metric & \multicolumn{2}{|c}{AUPRC (in \%)} & \multicolumn{2}{|c||}{AUROC (in \%)} & \multicolumn{2}{c}{Kappa ($\times 100$)} \\
\midrule

Head &          Linear &             MLP &          Linear &             MLP &          Linear &             MLP \\
\midrule
\midrule

Seq2-Seq-AE                  &  17.5 $\pm$ 1.3 &  19.3 $\pm$ 1.5 &  84.8 $\pm$ 0.6 &  86.8 $\pm$ 0.4 &  38.2 $\pm$ 0.6 &  41.6 $\pm$ 0.3 \\
Seq2Seq-AE-forecast         &  24.7 $\pm$ 1.3 &  28.7 $\pm$ 1.1 &  87.7 $\pm$ 0.4 &  89.7 $\pm$ 0.2 &  40.3 $\pm$ 0.3 &  42.2 $\pm$ 0.3 \\
CL                         &  31.0 $\pm$ 0.6 &  34.7 $\pm$ 0.4 &  88.3 $\pm$ 0.3 & 90.3 $\pm$ 0.2 &  40.4 $\pm$ 0.2 &  \textbf{43.2} $\pm$ 0.2 \\
SACL \citep{cheng2020subject} &  18.4 $\pm$ 1.9 &  29.3 $\pm$ 0.9 &  81.8 $\pm$ 1.3 &  87.5 $\pm$ 0.4 &  32.6 $\pm$ 2.0 &  40.1 $\pm$ 0.5 \\
CLOCS \citep{kiyasseh2020clocs}   & 29.9 $\pm$ 0.7 &  32.2 $\pm$ 0.8 &  89.5 $\pm$ 0.3 &  90.5 $\pm$ 0.2 &  41.7 $\pm$ 0.2 &  \textbf{43.0} $\pm$ 0.2 \\
$\NCLw$ (Ours)          &  31.2 $\pm$ 0.5 &  \textbf{35.1} $\pm$ 0.4 &  88.9 $\pm$ 0.3 &  \textbf{90.8} $\pm$ 0.2 &  40.5 $\pm$ 0.3 &  \textbf{43.2} $\pm$ 0.2 \\ 
\midrule
\midrule
End-to-End          &  34.3 $\pm$ 1.1 &  34.2 $\pm$ 0.6 & 90.6 $\pm$ 0.3 & 90.6 $\pm$ 0.2 &  43.3 $\pm$ 0.2 &  43.4 $\pm$ 0.2 \\

SCL (D) \citep{khosla2020supervised}      &  32.1 $\pm$ 0.9 &  31.9 $\pm$ 1.1 &  89.9 $\pm$ 0.3 &  89.5 $\pm$ 0.3 &  35.9 $\pm$ 0.7 &  40.2 $\pm$ 0.4 \\
SCL (L) \citep{khosla2020supervised}    &  30.6 $\pm$ 1.6 &  31.3 $\pm$ 0.8 &  86.1 $\pm$ 1.0 &  88.7 $\pm$ 0.4 &  41.3 $\pm$ 0.7 &  41.8 $\pm$ 0.4 \\
$\NCLY$ (D) (Ours)   &  \textbf{37.0} $\pm$ 0.6 &  \textbf{37.1} $\pm$ 0.7 &  90.3 $\pm$ 0.2 &  \textbf{90.9} $\pm$ 0.1 &  40.8 $\pm$ 0.3 &  43.3 $\pm$ 0.2 \\
$\NCLY$ (L) (Ours) &  33.5 $\pm$ 1.0 &  36.0 $\pm$ 0.6 &  88.2 $\pm$ 0.5 &  90.5 $\pm$ 0.2 &  \textbf{43.7} $\pm$ 0.2 &  \textbf{43.8} $\pm$ 0.3 \\
\bottomrule
\end{tabular}
\end{small}
\end{center}
\vskip -0.1in
\label{tab:mimic-iii}
\end{table*}

\begin{table*}[tbh!]
\caption{Results on the Physionet 2019 dataset. (Top rows) Unsupervised methods; (Bottom rows) Supervised methods.All scores are averaged over 20 runs such that the reported score is of the form $mean \pm std$. In bold are the methods within one standard deviation of best one for each setting. Evaluation metrics were scaled to 100 for readability purposes. }
\vskip 0.1in
\begin{center}
\begin{small}
\begin{tabular}{l|ll|ll|ll}
\toprule
Task & \multicolumn{6}{|c}{Sepsis onset prediction} \\
\midrule

Metric & \multicolumn{2}{|c}{AUPRC (in \%)} & \multicolumn{2}{|c|}{AUROC (in \%)} & \multicolumn{2}{c}{Utility ($\times 100$)} \\
\midrule

Head &          Linear &             MLP &          Linear &             MLP &          Linear &             MLP \\
\midrule
\midrule

Seq2-Seq-AE          &  7.0 $\pm$ 0.3 &  7.8 $\pm$ 0.4 &  77.1 $\pm$ 0.5 &  78.1 $\pm$ 0.6 &  26.8 $\pm$ 1.0 &  27.2 $\pm$ 1.0 \\
Seq2-Seq-AE-forecast &  6.6 $\pm$ 0.3 &  7.3 $\pm$ 0.3 &  75.8 $\pm$ 0.9 &  76.9 $\pm$ 0.5 &  23.5 $\pm$ 1.5 &  23.8 $\pm$ 1.2 \\
CL                   &   7.9 $\pm$ 0.4 &   \textbf{9.5} $\pm$ 0.4 &  78.2 $\pm$ 0.3 &  80.2 $\pm$ 0.4 &  26.2 $\pm$ 0.8 &  \textbf{29.7} $\pm$ 1.0 \\
SACL   \citep{cheng2020subject}&   6.5 $\pm$ 0.3 &   7.6 $\pm$ 0.3 &  73.0 $\pm$ 1.2 &  75.3 $\pm$ 0.8 &  20.5 $\pm$ 2.5 &  24.2 $\pm$ 1.1 \\
CLOCS   \citep{kiyasseh2020clocs}    &   7.1 $\pm$ 0.5 &   7.3 $\pm$ 0.4 &  77.2 $\pm$ 0.5 &  78.8 $\pm$ 0.4 &  23.0 $\pm$ 1.1 &  25.8 $\pm$ 0.9 \\
$\NCLw$ (Ours)     &   8.2 $\pm$ 0.4 &   \textbf{9.3} $\pm$ 0.5 &  78.8 $\pm$ 0.3 &  \textbf{80.7} $\pm$ 0.3 &  27.2 $\pm$ 1.0 &  \textbf{30.2} $\pm$ 1.0 \\
\midrule
\midrule
End-to-End  &  7.6 $\pm$ 0.2 &  8.1 $\pm$ 0.4 &  78.9 $\pm$ 0.3 & 78.8 $\pm$ 0.4 &  27.9 $\pm$ 0.8 &  27.5 $\pm$ 1.0 \\
SCL \citep{khosla2020supervised}         &   6.7 $\pm$ 0.6 &   6.0 $\pm$ 0.5 &  73.1 $\pm$ 1.7 &  70.0 $\pm$ 1.9 &  20.2 $\pm$ 2.7 &  20.6 $\pm$ 1.7 \\
$\NCLY$  (Ours) &  \textbf{10.0} $\pm$ 0.5 &  \textbf{10.1} $\pm$ 0.3 &  80.3 $\pm$ 0.4 &  \textbf{80.8} $\pm$ 0.2 &  \textbf{32.6} $\pm$ 1.0 &  \textbf{31.9} $\pm$ 0.9 \\
\bottomrule
\end{tabular}
\end{small}
\end{center}
\vskip -0.1in
\label{tab:physionet2019}
\end{table*}

\begin{table*}[t]
\caption{Results for the limited-labeled data scenario on the Decompensation task. Each method was run on 4 random splits of the training data. Splitting was done at a patient level and splits were stratified to preserve the label prevalence.}
\vskip 0.1in
\begin{center}
\begin{small}
\begin{tabular}{l|ll|ll|ll}
\toprule
Task & \multicolumn{6}{|c}{Decompensation} \\
\midrule
Labels & \multicolumn{2}{|c}{1 \%} & \multicolumn{2}{|c|}{10 \%} & \multicolumn{2}{c}{50 \%} \\
\midrule
Metric & AUPRC (in \%) &  AUROC (in \%) & AUPRC (in \%) &   AUROC (in \%) &          AUPRC (in \%) &  AUROC (in \%)  \\
\midrule
Seq2-Seq-AE          &   8.3 $\pm$ 1.4 &  79.4 $\pm$ 1.7 &  14.4 $\pm$ 1.0 &  84.7 $\pm$ 0.6 &  17.5 $\pm$ 1.3 &  86.1 $\pm$ 0.5 \\
Seq2-Seq-AE-forecast &  12.1 $\pm$ 1.5 &  83.0 $\pm$ 1.3 &  21.7 $\pm$ 2.0 &  87.7 $\pm$ 0.4 &  26.5 $\pm$ 1.4 &  89.1 $\pm$ 0.3 \\
CL         &  \textbf{21.6} $\pm$ 4.0 &  \textbf{84.9} $\pm$ 1.0 &  \textbf{30.6} $\pm$ 0.9 &  88.6 $\pm$ 0.3 &  \textbf{33.9} $\pm$ 0.4 &  89.8 $\pm$ 0.2 \\
SACL       &  12.3 $\pm$ 3.0 &  77.1 $\pm$ 1.6 &  20.6 $\pm$ 1.5 &  83.6 $\pm$ 0.8 &  27.4 $\pm$ 1.1 &  86.6 $\pm$ 0.5 \\
CLOCS      &  13.1 $\pm$ 3.3 &  83.7 $\pm$ 1.7 &  26.1 $\pm$ 1.8 &  \textbf{88.9} $\pm$ 0.4 &  31.2 $\pm$ 0.9 &  \textbf{90.1} $\pm$ 0.2 \\
$\NCLw$  (Ours)    &  19.8 $\pm$ 3.5 &  \textbf{85.2} $\pm$ 1.3 &  29.8 $\pm$ 1.3 &  \textbf{89.1} $\pm$ 0.3 &  \textbf{34.1} $\pm$ 0.6 &  \textbf{90.3} $\pm$ 0.2 \\
End-to-End &  13.1 $\pm$ 2.9 &  82.6 $\pm$ 1.4 &  25.8 $\pm$ 1.5 &  88.2 $\pm$ 0.4 &  32.3 $\pm$ 0.6 &  89.9 $\pm$ 0.2 \\
\bottomrule
\end{tabular}
\end{small}
\end{center}
\vskip -0.1in
\label{tab:fl}
\end{table*}

\paragraph{$\NCLw$ closes the gap to end-to-end training.}
In the unsupervised setting, from Table \ref{tab:mimic-iii}, we observe that $\NCLw$, with an MLP head, is the only method to beat end-to-end training on both metrics for the \textit{Decompensation} task. Also, by performing on par with the best unsupervised methods on \textit{Length-of-stay} predictions, despite its additional parameters, $\NCLw$ is not task-specific. Moreover, as shown in Figure \ref{fig:perf_trade-off}, using $n_w$ alone is not sufficient. When not coupled to $\ltotal$, performance significantly decreases.

While CLOCS exhibits good performance on the MIMIC-III Benchmark, it fails to do so for \textit{Sepsis} predictions (Table \ref{tab:physionet2019}). Furthermore, for \textit{Sepsis} predictions, among the contrastive methods designed to tackle patient-induced heterogeneity, $\NCLw$ is the only one that improves performance over end-to-end training. However, CL achieving similar performances as our method on the task suggests some limitations inherent to $n_w$.

\paragraph{$\NCLY$ significantly improves over SCL.}
In the supervised setting, SCL has shown to be highly unstable to train. As shown in Table \ref{tab:mimic-iii} and \ref{tab:physionet2019}, SCL fails to learn a good representation. It has the lowest performance among all methods, both supervised and unsupervised.

We believe the low performance of SCL is due to highly similar time-series segments that can have different labels. It results in an excessively challenging pretext task known to hurt downstream performance \citep{tian2020makes}. As shown in Figure \ref{fig:perf_trade-off-ncly}, easing the pretext task, by lowering $\alpha$, increases the model contrastive accuracy. On the downstream task, it yields a similar trade-off in performance as \citet{tian2020makes} in the unsupervised case.

When using our objective $\ltotal$ with the same neighborhood function $n_Y$ as SCL, we observe a significant improvement on all tasks. On the MIMIC-III Benchmark (Table \ref{tab:mimic-iii}), when trained with the task labels, it outperforms all other methods, including SCL. More importantly, the features learned by $\NCLY$ transfer much better to the other task. Even when trained with \textit{Length-of-stay} labels, for the \textit{Decompensation} task, our method still outperforms end-to-end training and all instances of SCL. Finally, as shown in Table \ref{tab:physionet2019}, while SCL fails on the sepsis task, $\NCLY$ surpasses all other methods.

\paragraph{Frozen linear evaluation is limited for patient state representation. } \label{sec:mlp_vs_linear}

A common way to compare representation learning approaches in the literature has been to use a linear classifier on top of a learned representation. In our work, we show (Table \ref{tab:mimic-iii}, Table \ref{tab:physionet2019}) that a non-linear classifier yields significantly better performances for all representation learning methods. However, end-to-end training does not benefit from a non-linear head. More than not capturing the full potential of a representation, linear evaluation can even be misleading. For instance, under linear evaluation, CLOCS outperforms our methods in AUROC for \textit{Decompensation}. However, with a MLP head, even if both improve, their relative ordering in performance changes.

\paragraph{Reducing amount of labels improves over supervised training.}
To evaluate unsupervised representations, it is common practice to compare them to their end-to-end counterparts while decreasing the amount of labeled data. To this end, we report results using fractions of \textit{Decompensation} label amounts in Table \ref{tab:fl}.  We show that $\NCLw$ significantly outperforms end-to-end training and other patient-designed methods on both AUROC and AUPRC when the number of labeled patients is reduced. However, while our method stays competitive with regular CL in AUROC, it is not the case for AUPRC. Such a discrepancy between the two metrics suggests an increasing proportion of false positives when reducing labeled data for $\NCLw$.

\section{Conclusion}

In this paper, our aim is to bring state-of-the-art contrastive learning methods one step closer to being applied to online patient state monitoring in the ICU. Our work addresses the domain heterogeneity problems inherent to this task by introducing a contrastive objective which encourages patient state representations to follow distributional assumptions dictated by prior knowledge.

By operating in conjunction with existing augmentation techniques, we are able to make contrastive learning a more expressive framework for working with challenging real-world datasets and incorporating adjacent information and prior knowledge.

When fully-unsupervised, our method shows competitive results over end-to-end training and previous self-supervised approaches for biosignals. Besides, when used in a supervised manner, our framework considerably improves over existing supervised contrastive learning methods.

In addition to the choice of data augmentations, our method depends strongly on the definition of the neighborhood function. Our binary neighborhood framework is the first step in encouraging higher-order behavior of contrastive learning representations via loss functions. We believe that relaxing this definition from being binary to categorical or continuous is a promising direction for future work.

 \section{Acknowledgements}
This work is supported by the Grant No. 205321\_176005 of the Swiss National Science Foundation and ETH core funding (to Gunnar Rätsch). In addition, Francesco Locatello is supported by the Max Planck ETH Center for Learning Systems and by a Google Ph.D Fellowship.  
\bibliography{bibliography}
\bibliographystyle{icml2021}

\newpage

\appendix

\onecolumn

\section{Architecture} \label{appendix_arch}
\begin{figure*}[th!]
    \centering
    \includegraphics[width=\columnwidth]{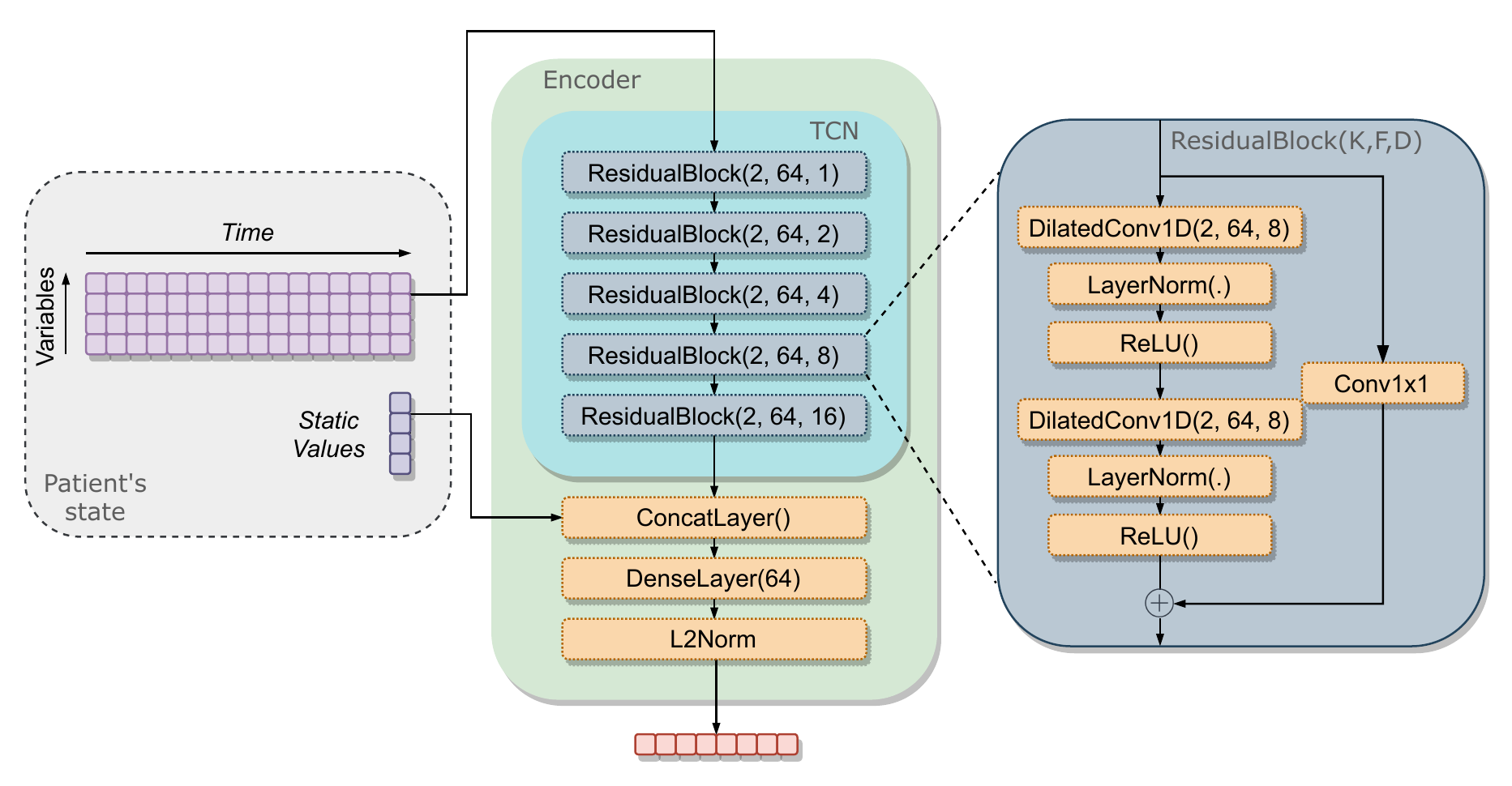}

    \caption{Encoder architecture we used for all methods. In the figure, K, F, and D represent respectively, the kernel size, number of filters, and dilation rate. We use a similar TCN block than the original paper \citep{bai1803empirical} with the exception that we use layer normalization. We use a fully-connected layer to incorporate static features in the representation. Finally, we normalize this representation to the unit sphere as in \citet{he2020momentum}}
    \label{fig:architeture}
\end{figure*}

In this section, we expand on the details of our architecture. The full architecture of the encoder is depicted in Figure \ref{fig:architeture}. For the non-linear projector and classifier, we used an inner layer with the same dimension as the representation size. Thus for all tasks, we used an inner dimension of 64. Because we deal with time-series, we used causal dilated convolutions to not break temporal ordering. We built our pipeline using \texttt{tensorflow 2.3} and \texttt{keras-tcn 3.3}.

\FloatBarrier

\section{Data augmentation} \label{appendix_data}
In this section, we further expand on the data augmentations used in all contrastive methods. To choose each function's hyperparameters we performed a random search on the validation performance for both MIMIC-III Benchmark and Physionet 2019 for the regular CL method. As a result of the random search, we chose the following parameters. 

\begin{enumerate}
    \item \textbf{History Crop}: We apply a crop with a probability of $0.5$ and minimum size of $50\%$ of the initial sequence.
    \item \textbf{History Cutout}: We apply time cutout of $8$ steps with a probability of $0.8$.
    \item \textbf{Channel Dropout}: We mask out each channel randomly with a probability of $0.2$
    \item \textbf{Gaussian Noise}: We add random Gaussian noise to each variable independently with a standard deviation of $0.1$
\end{enumerate}

\begin{figure}[pth!]
\begin{center}
\centerline{\includegraphics[scale=0.8]{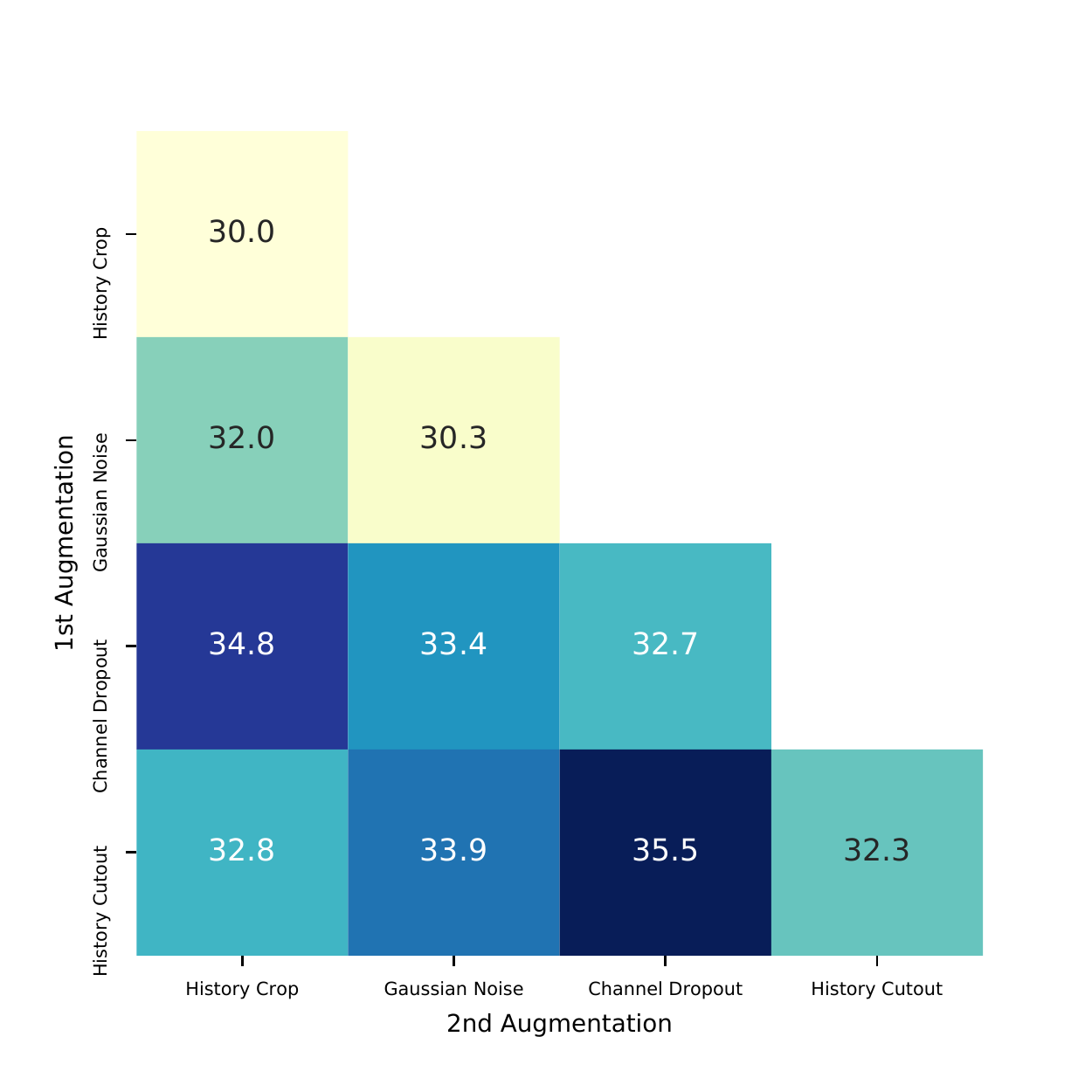}}
\caption{Comparison of performance between different choices of augmentation. Result are reported on validation set AUPRC for the decompensation task and are averaged over 5 seeds. }
\label{fig:da}
\end{center}
\end{figure}

Also, we verify that composing augmentations \citep{chen2020simple} improves performances. We find, as in \citet{cheng2020subject} and \cite{kiyasseh2020clocs}, that composing temporal and spatial augmentations yields the best performances as shown in Figure \ref{fig:da}. It obtains lower performance than composing all transformations, which achieves an AUPRC of $35.5$ on validation for the 5 same seeds. Therefore, we applied these four augmentations sequentially to both branches of the pipeline for all contrastive methods.

\section{Data sets} \label{appendix_ds}
In this section we expand further on the datasets we performed experiments on. 
\subsection{MIMIC-III Benchmark}
As shown in Table \ref{tab:variables_MIMIC}, MIMIC-III Benchmark provides 17 measurements in addition to the time since admission. After one-hot encoding of the categorical features, we obtain an input dimension of 42. 

In Table \ref{tab:data-set-desc-mimic-iii}, we detail the splitting and prevalence of the dataset. We observe that, compared to Physionet 2019, the length of patient stays are significantly greater. Moreover, we also observe that decompensation is a highly unbalanced task. 

\begin{figure}[pth!]
\begin{center}
\centerline{\includegraphics[scale=0.8]{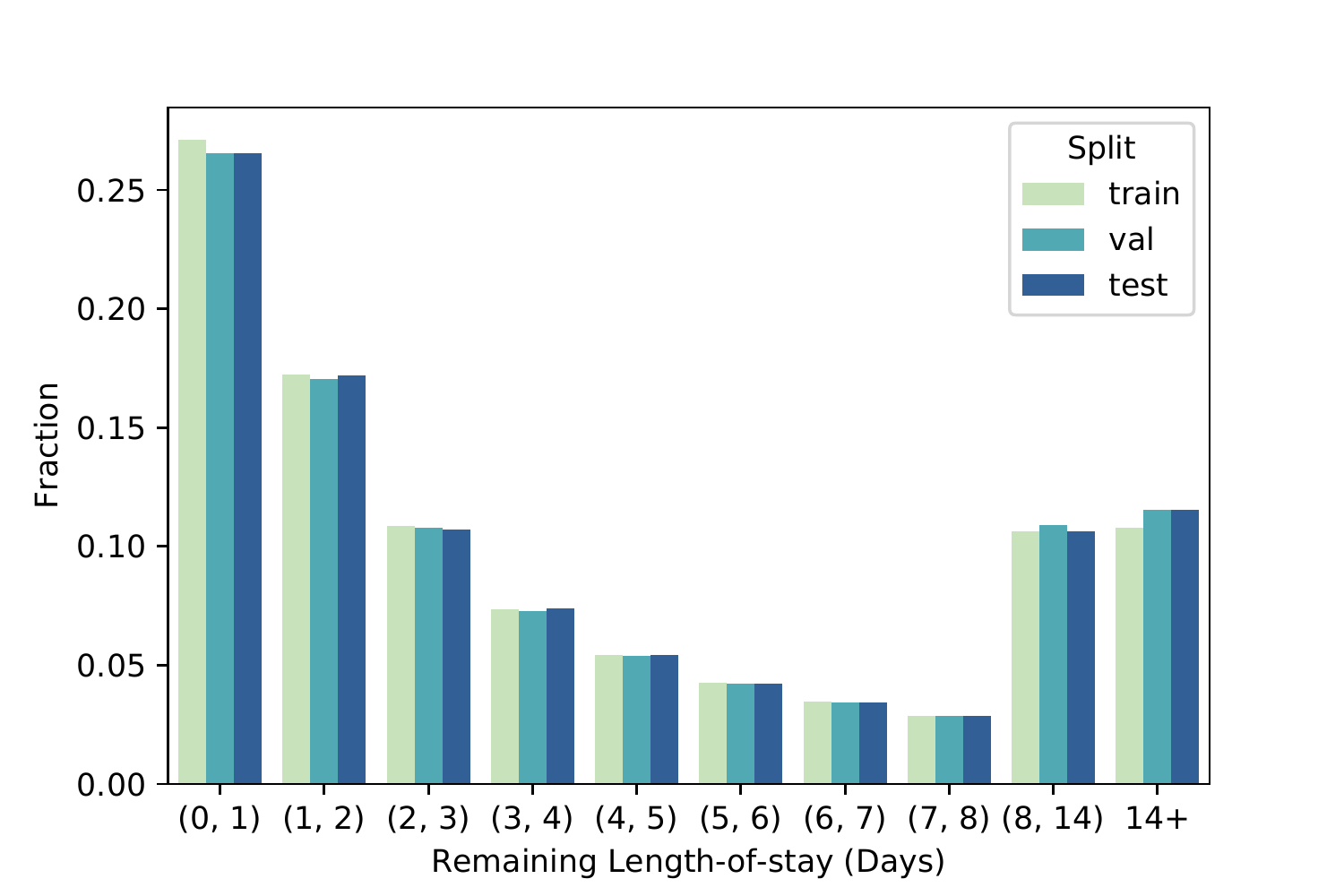}}
\vskip -0.1in
\caption{Prevalence of each temporal bin used in the \textit{Length-of-stay} task. We used the same bins as \cite{harutyunyan2019multitask}.}
\label{fig:fraction}
\end{center}
\end{figure}

\begin{table}[h!]
\caption{Number of patients and samples in the full 
 data-set as well as for individual predictive tasks.}
\vskip 0.1in
\centering
\begin{tabular}{llll}
\toprule
 \textbf{MIMIC-III} & \multicolumn{3}{c}{\textbf{Number of patients}} \\
 & Train & Test & Val \\
 & 29250 & 6281 & 6371 \\
\midrule
\textbf{Length of stay} & \multicolumn{3}{c}{\textbf{Number of samples}} \\
 & Train & Test & Val \\
 & 2,586,619 & 563,742 & 572,032 \\
 \midrule
\textbf{Decompensation} & \multicolumn{3}{c}{\textbf{Number of samples}}  \\
& Train & Test & Val \\
 & 2,377,738 & 523,200 & 530,638 \\
 \midrule
& \multicolumn{3}{c}{\textbf{Number of positives}} \\
 & Train & Test & Val \\
 & 49,260 & 9,683 & 11,752 \\
\bottomrule
\end{tabular}

\label{tab:data-set-desc-mimic-iii}
\end{table}

\begin{table}[pth]
\caption{Measurements recorded and re-sampled hourly in the MIMIC-III benchmark dataset. BP: Blood pressure, MAP: Mean arterial pressure, FiO$_2$: Fraction of inspired oxygen. GCS: Glasgow Coma Scale. SpO$_2$: Pulse oxygen saturation.}
\vskip 0.1in
    \centering
\begin{tabular}{ll}
\toprule
                        \textbf{Measurement} &        \textbf{Type} \\
\midrule
            Time since admission  &  Continuous \\
                             Height &  Static (Continuous) \\
              Capillary refill rate &  Categorical \\
     GCS eye opening &  Categorical \\
  GCS motor response &  Categorical \\
  GCS verbal response &  Categorical \\
           GCS total &  Categorical \\
            Diastolic BP &  Continuous \\
           FiO$_2$ &  Continuous \\
                            Glucose &  Continuous \\
                         Heart Rate &  Continuous \\
                MAP &  Continuous \\
                  SpO$_2$ &  Continuous \\
                   Respiratory rate &  Continuous \\
            Systolic BP &  Continuous \\
                        Temperature &  Continuous \\
                             Weight &  Continuous \\
                                 pH &  Continuous \\
\bottomrule
\end{tabular}
    \label{tab:variables_MIMIC}
\end{table}

\FloatBarrier

\subsection{Physionet 2019}
As shown in Table \ref{tab:Variables Physionet} Physionet 2019 provides 40 measurements. As all categorical features are binary, the final input dimension is 40 as well. In Table \ref{tab:data-set-desc-physionet2019}, we detail the splitting and prevalence of the dataset. We, once again, highlight the very low prevalence of positive labels.

\begin{table}[th!]
\caption{Measurement recorded and re-sampled hourly for Physionet 2019. BP: Blood pressure, MAP: Mean arterial pressure, FiO$_2$: Fraction of inspired oxygen. PaCO$_2$: Partial pressure of carbon dioxide from arterial blood. SaO$_2$: Oxygen saturation from arterial blood. SpO$_2$: Pulse oxygen saturation. }
\vskip  0.1in
    \centering
\begin{tabular}{ll}
\toprule
     Measurement &        Type \\
\midrule
             Time since admission (ICU) &  Continuous \\
             Age &  Static (Continuous) \\
            Gender  &  Static (Categorical) \\
     Hospital Admission Time &  Static (Continuous) \\
         ICU Unit 1 &  Static (Categorical) \\
          ICU Unit 2 &  Static (Categorical) \\
          
          Heart rate & Continuous \\
          SpO$_2$ & Continuous \\
          Temperature & Continuous \\
          Systolic BP & Continuous \\
          MAP & Continuous \\
          Diastolic BP & Continuous \\
          Respiratory rate & Continuous \\
          End tidal carbon dioxide & Continuous \\
          Excess Bicarbonate &  Continuous \\
          Bicarbonate &  Continuous \\
          FiO$_2$ &  Continuous \\
          PaCO$_2$ &  Continuous \\

\bottomrule
\end{tabular}
\quad
\begin{tabular}{ll}
\toprule
     Measurement &        Type \\
\midrule

          SaO$_2$ &  Continuous \\
          Aspartate transaminase &  Continuous \\
          Blood urea nitrogen &  Continuous \\
          Alkaline phosphatase &  Continuous \\
          Calcium &  Continuous \\
          Chloride &  Continuous \\
          Creatinine &  Continuous \\
          Bilirubin direct &  Continuous \\
          Total bilirubin &  Continuous \\
          Serum glucose &  Continuous \\
          Lactic acid &  Continuous \\
          Troponin I &  Continuous \\
          Hematocrit &  Continuous \\
          Hemoglobin &  Continuous \\
          Partial Thromboplastin time &  Continuous \\
          Leukocyte count &  Continuous \\
          Fibrinogen &  Continuous \\
          Platelets &  Continuous \\
\bottomrule
\end{tabular}

    \label{tab:Variables Physionet}

\end{table}

\begin{table}[ht!]
\caption{Description of Physionet 2019 statistics by patient and sample.}
\vskip  0.1in
\centering
\begin{tabular}{llll}
\toprule
 \textbf{Physionet 2019} & \multicolumn{3}{c}{\textbf{Number of patients}} \\
 & Train & Test & Val \\
 & 25,813 & 8,066 & 6,454 \\
\midrule
\textbf{Sepsis onset} & \multicolumn{3}{c}{\textbf{Number of samples}}  \\
& Train & Test & Val \\
 & 992,732 & 312,078 & 247,283 \\
 \midrule
& \multicolumn{3}{c}{\textbf{Number of positives}} \\
 & Train & Test & Val \\
 & 17,891 & 5,550 & 4,475 \\
\bottomrule
\end{tabular}

\label{tab:data-set-desc-physionet2019}
\end{table}
\FloatBarrier

\newpage

\section{Hyperparameter selection}\label{hp_appendix}

 We tuned all existing hyperparameters over validation performances. For MIMIC-III, 
 we used AUPRC on \textit{Decompensation} task as a reference. For Physionet 2019 we used the
 Utility metric from \cite{reyna2019early}.
 
\subsection{Architecture parameters}
The main hyperparameters of the TCN architecture are the kernel size and the number of filters. We
tuned these parameters on the End-to-end model and then used them for all other methods. 

\begin{table}[pth]
\centering
\begin{tabular}{c|c}
\toprule
Kernel Size &        AUPRC (Validation set) \\
\midrule
2         &  \textbf{37.0} $\pm$ 0.5 \\
4         &  36.7 $\pm$ 0.4 \\
8         &  35.7 $\pm$ 0.6 \\
\bottomrule
\end{tabular}
\quad
\begin{tabular}{c|c}
\toprule
Number of filters &        AUPRC (Validation set) \\
\midrule
16        &  37.1 $\pm$ 0.4 \\
32        &  37.0 $\pm$ 0.5 \\
64        &  \textbf{37.3} $\pm$ 0.5 \\
128       &  37.1 $\pm$ 0.5 \\
256       &  36.8 $\pm$ 1.0 \\
512       &  35.7 $\pm$ 2.0 \\
\bottomrule
\end{tabular}
\caption{(a) Impact of the kernel size parameter on the validation AUPRC metric for end-to-end training on MIMIC-III decompensation task. Results are averaged over 5 seeds and number of filters was set to 32, (b) Impact of the number of filters on the validation AUPRC metric for end-to-end training on the MIMIC-III decompensation task. Results are averaged over 5 seeds and kernel size was set to 2}
\end{table}

\begin{table}[h]
\centering
\begin{tabular}{c|c}
\toprule
Kernel Size &        Utility (Validation set) \\
\midrule
2         &  \textbf{28.7} $\pm$ 0.7 \\
4         &  28.0 $\pm$ 1.1 \\
8         &  \textbf{29.1 $\pm$ 1.6} \\
\bottomrule
\end{tabular}
\quad
\begin{tabular}{c|c}
\toprule
Number of filters &        Utility (Validation set) \\
\midrule
16        &  27.8 $\pm$ 1.4 \\
32        &  28.8 $\pm$ 0.6 \\
64        &  \textbf{29.0} $\pm$ 0.8 \\
128       &  28.8 $\pm$ 1.3 \\
256       &  26.8 $\pm$ 3.1 \\
\bottomrule
\end{tabular}
\caption{(a) Impact of the kernel size parameter on the validation Utility metric for end-to-end training on Physionet 2019. Results are averaged over 5 seeds and the number of filters was set to 32., (b) Impact of the number of filter on the validation Utility metric for end-to-end training on Physionet 2019. Results are averaged over 5 seeds and the kernel size was set to 2}
\end{table}

\newpage
\subsection{Contrastive parameters}

\begin{figure}[th!]
    \centering
    \includegraphics[scale=0.6]{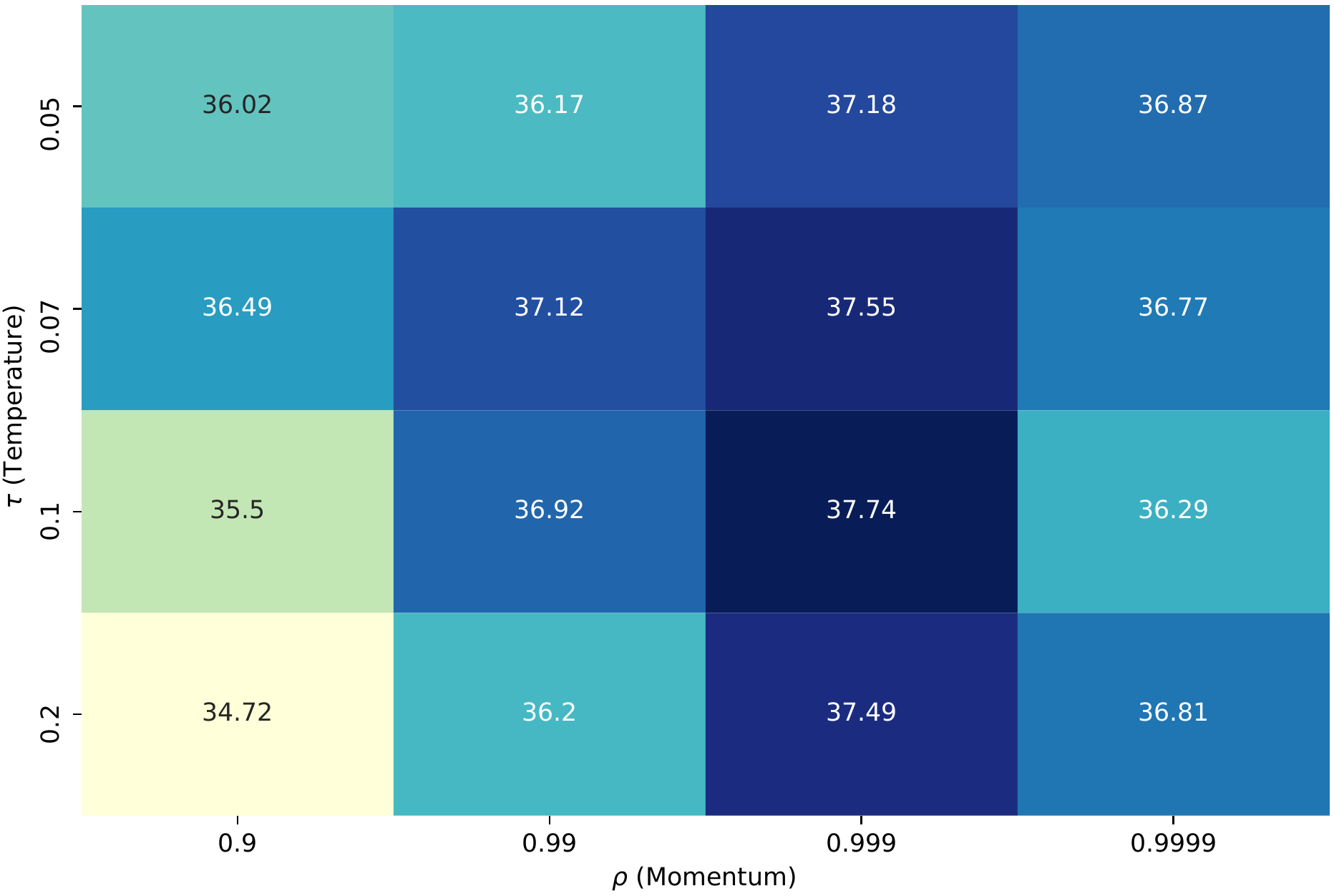}
    \caption{Grid search over $\tau$ (temperature) and $\rho$ (momentum) for regular Contrastive Learning method on MIMIC-III. Here result are averaged over 5 runs. Reported metric is AUPRC on validation set for \textit{Decompensation} task.}
    \label{fig:mimic_gs_t_m}
\end{figure}

\begin{figure}[th!]
    \centering
    \includegraphics[scale=0.6]{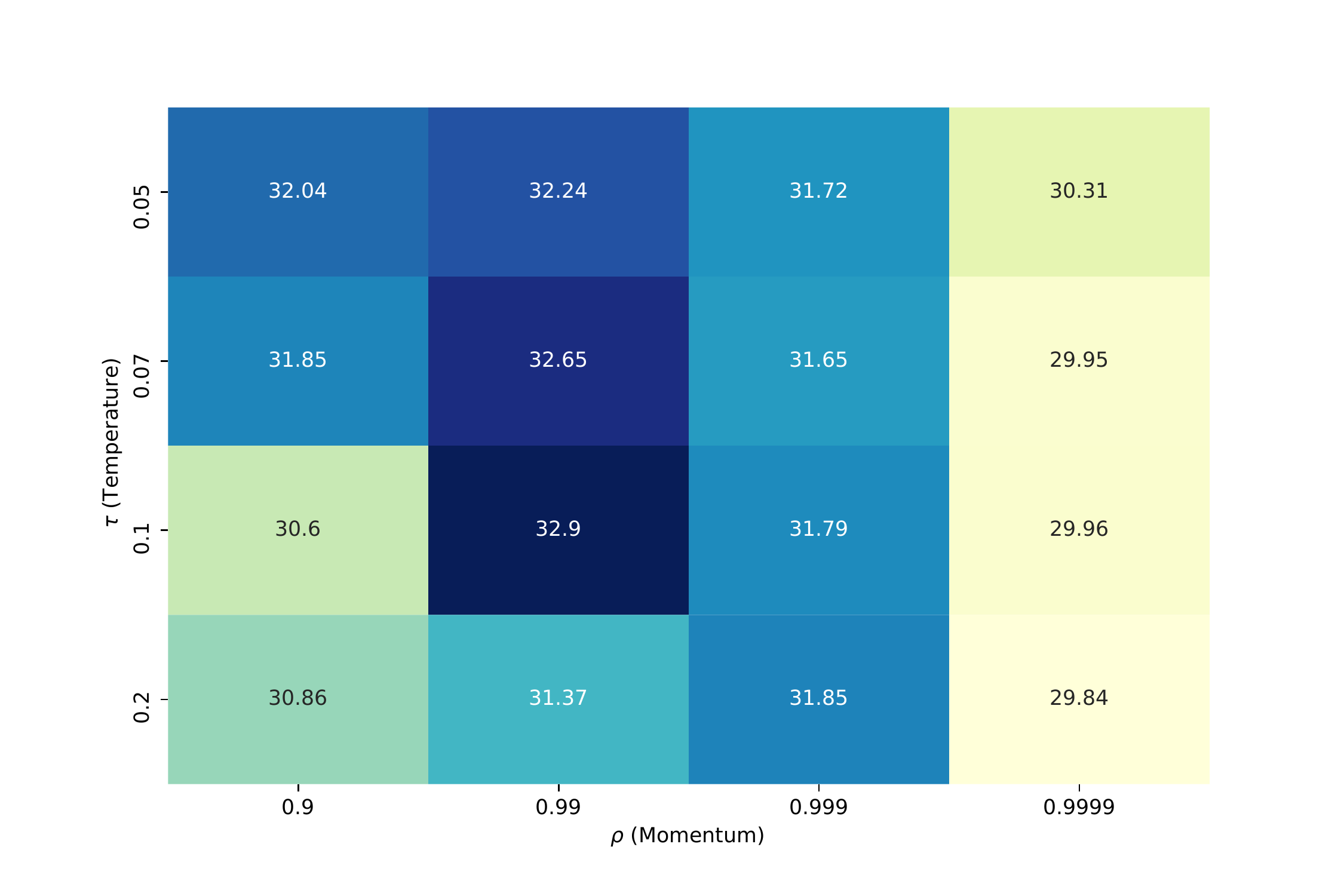}
    \caption{Grid search over $\tau$ (temperature) and $\rho$ (momentum) for regular Contrastive Learning method on Physionet 2019. Here result are averaged over 5 runs. Reported metric is Utility on validation set for sepsis task.}
    \label{fig:physionet_gs_t_m}
\end{figure}

The two main contrastive parameters shared across methods are the momentum $\rho$ and the temperature $\tau$. For a fair comparison, we used the same values for these parameters based on the performance of regular CL as shown in Figure \ref{fig:mimic_gs_t_m} and \ref{fig:physionet_gs_t_m}.

\subsection{Neighborhood parameters}
\begin{figure}[h]
    \centering
    \includegraphics[scale=0.5]{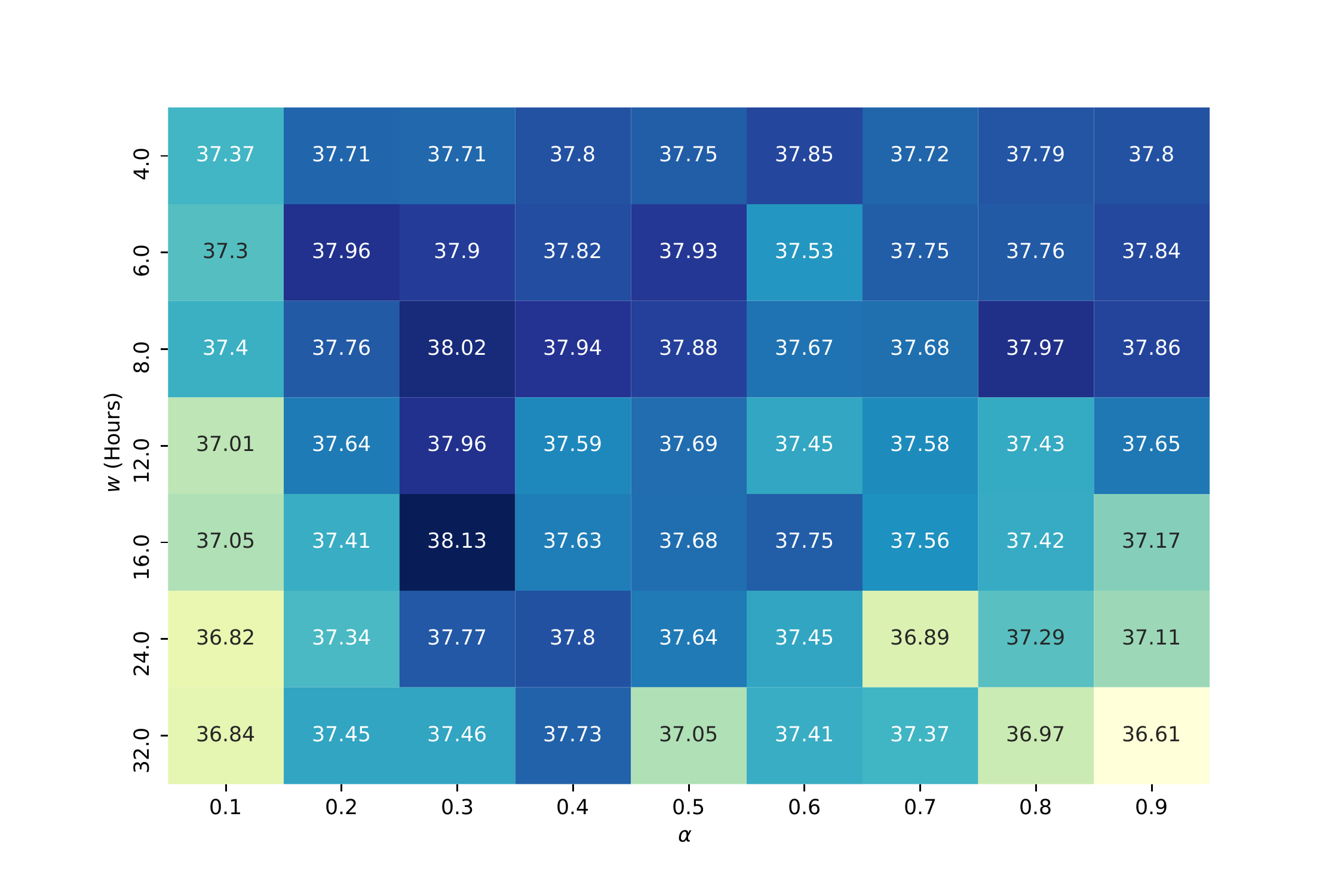}
    \caption{Grid search over $w$ and $\alpha$ for $\NCLw$ method on MIMIC-III Benchmark. Here result are averaged over 5 runs. Reported metric is AUPRC on validation set for decompensation task.}
    \label{fig:mimic_gs}
\end{figure}

\begin{figure}[h]
    \centering
    \includegraphics[scale=0.5]{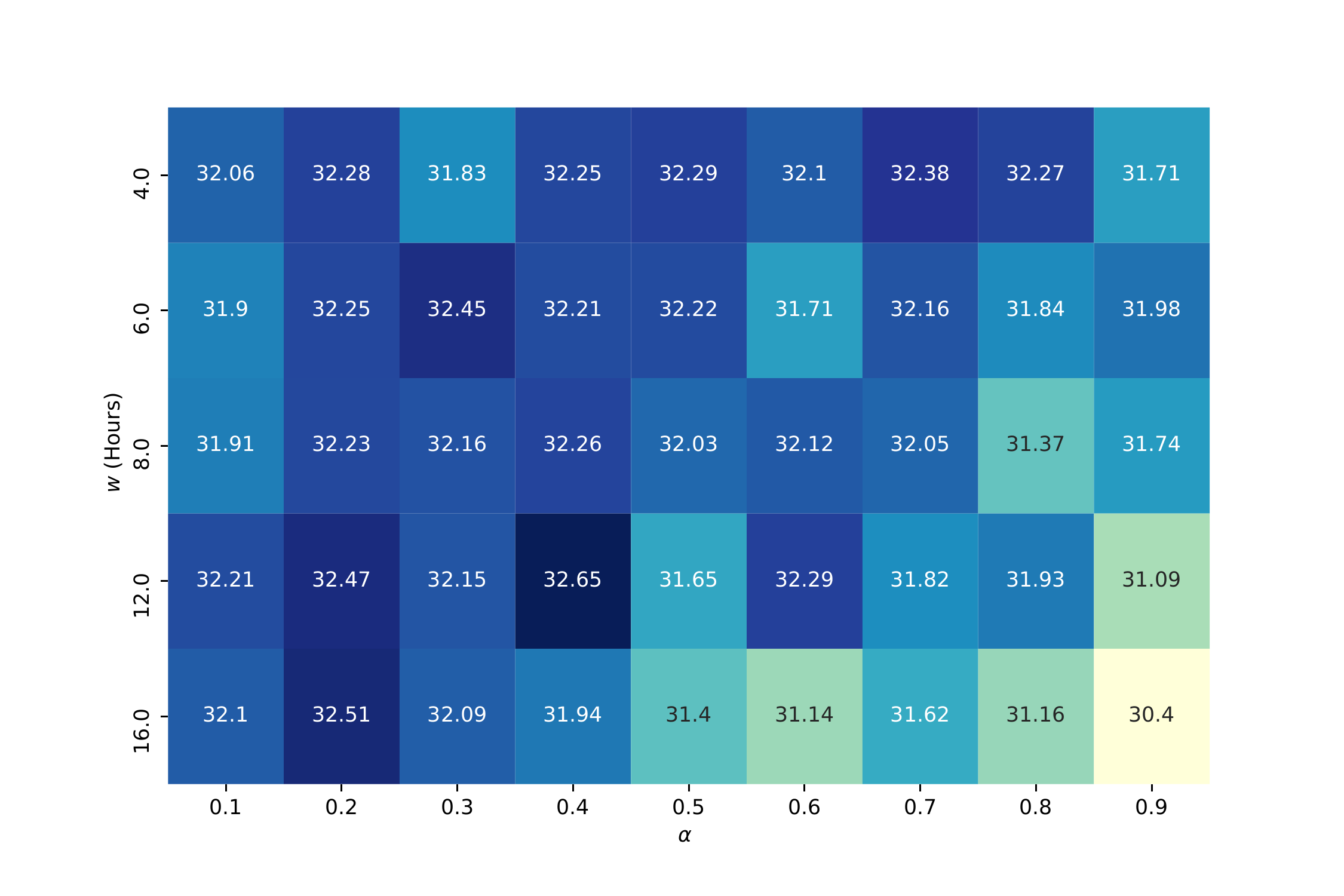}
    \caption{Grid search over $w$ and $\alpha$ for $\NCLw$ method on Physionet 2019. Here result are averaged over 5 runs. Reported metric is Utility on validation set for sepsis task.}
    \label{fig:physionet_gs}
\end{figure}

As shown in Figures \ref{fig:mimic_gs} and \ref{fig:physionet_gs}, we select the specific parameters to $n_w$ with a grid search over 5 runs. If parameters yielding good performance are stable for MIMIC-III, we found that performance on Utility metric varied significantly for Physionet 2019. We believe a reason for that is the fact these metrics depend on a threshold for making a prediction. Thus, contrary to AUROC or AUPRC, in addition to evaluating the model performances, it also evaluates its calibration.

\begin{figure}[h]
    \centering
    \includegraphics[width=\columnwidth]{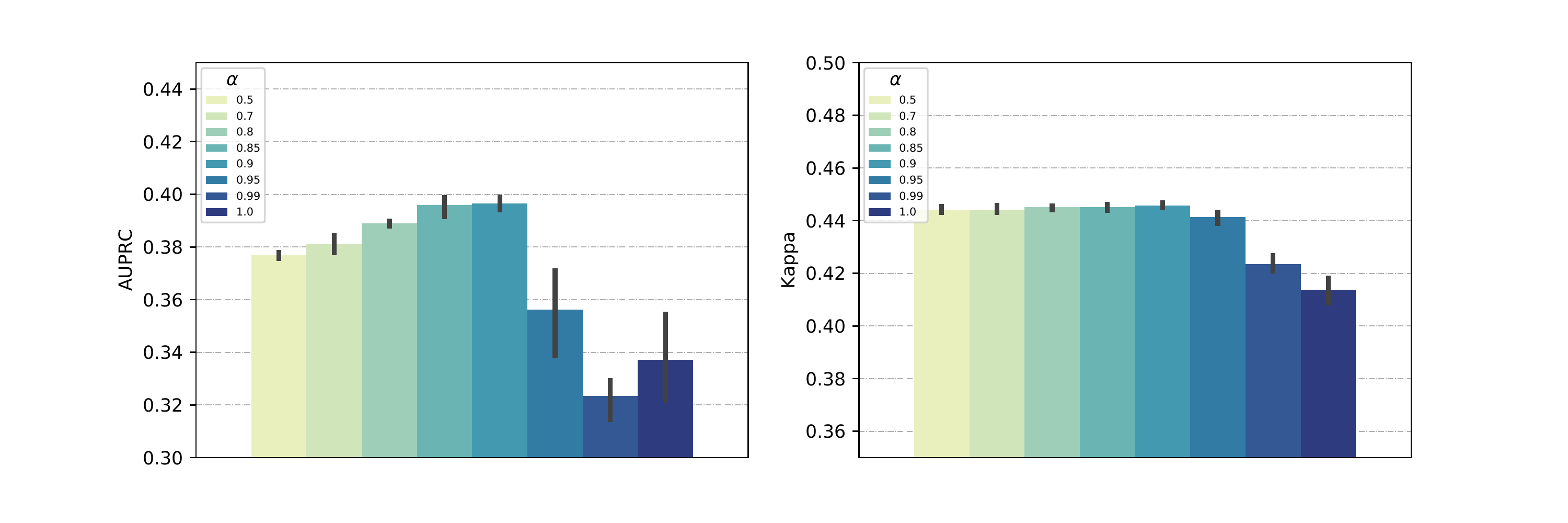}
    \caption{Parameter search over $\alpha$ for $\NCLY$ method on MIMIC-III Benchmark when trained using \textit{Decompensation} labels. Here result are averaged over 5 runs. Reported metric is AUPRC (for \textit{Decompensation}) and Kappa (for \textit{Length-of-stay})on validation set.}
    \label{fig:decomp_NCLY}
\end{figure}

\begin{figure}[h]
    \centering
    \includegraphics[width=\columnwidth]{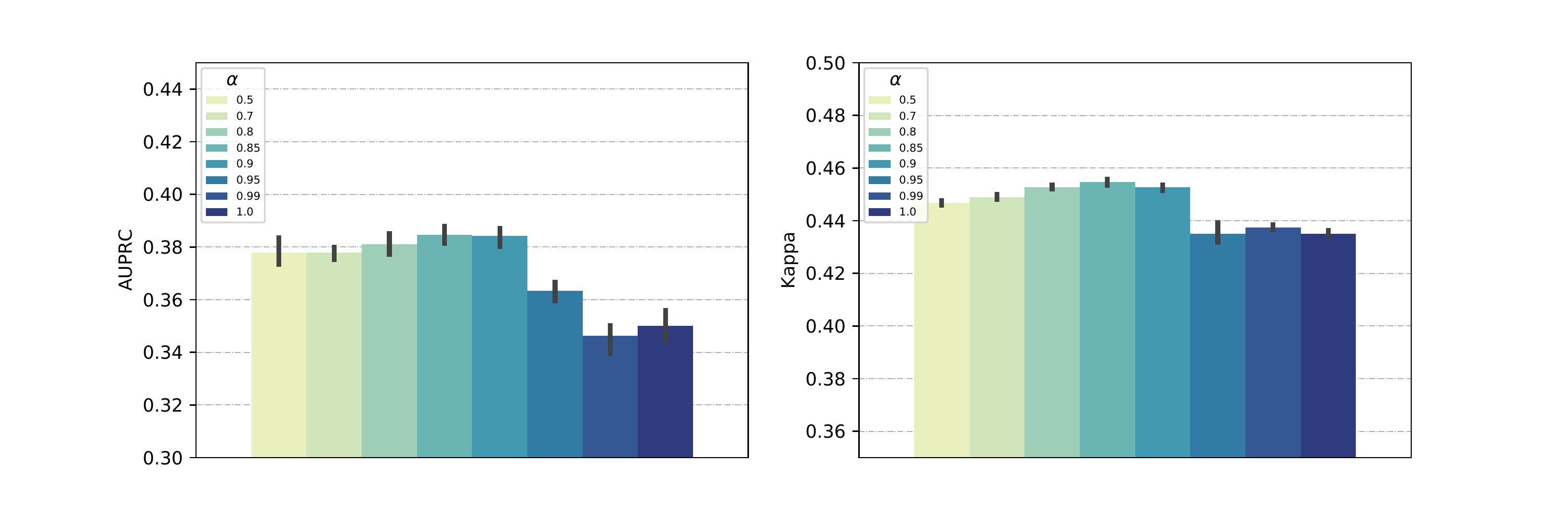}
    \caption{Parameter search over $\alpha$ for $\NCLY$ method on MIMIC-III Benchmark when trained using \textit{Length-of-stay} labels. Here result are averaged over 5 runs. Reported metric is AUPRC (for \textit{Decompensation}) and Kappa (for \textit{Length-of-stay})on validation set.}
    \label{fig:los_NCLY}
\end{figure}

\begin{figure}[h]
    \centering
    \includegraphics[scale=0.5]{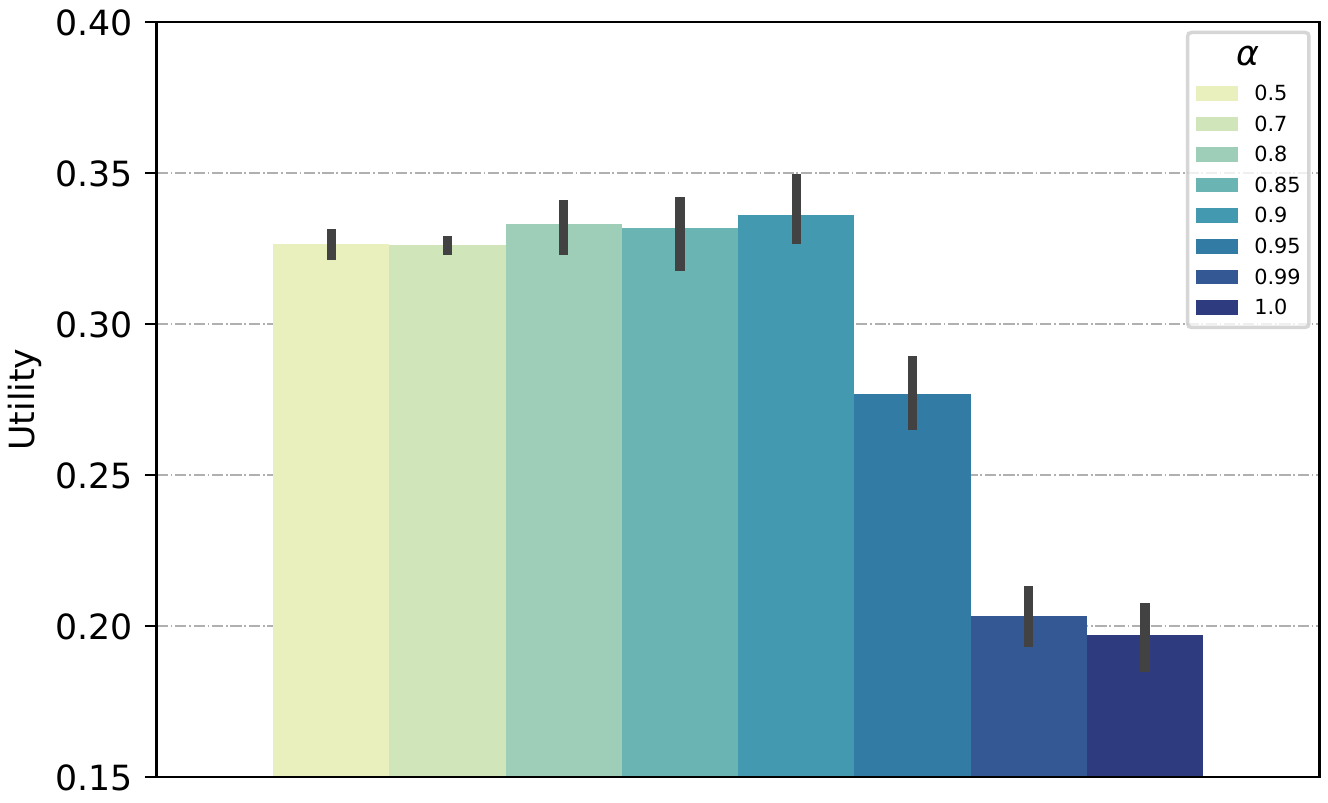}
    \caption{Parameter search over $\alpha$ for $\NCLY$ method on Physionet 2019. Here result are averaged over 5 runs. Reported metric is Utility on validation set for sepsis task.}
    \label{fig:sepsis_NCLY}
\end{figure}

As showed in Figures \ref{fig:decomp_NCLY},  \ref{fig:los_NCLY} and  \ref{fig:sepsis_NCLY} we selected $\alpha$ for $\NCLY$ on validation set performance. We observe that for all tasks, taking $\alpha=0.9$ yields the best performance on the training task and in transfer learning. 

\clearpage
\section{Supplementary Results}
\subsection{Semi-supervised neighborhood}
We explored another neighborhood possibility for our supervised approach, as the intersection of $N_w$ and $N_Y$, called $\NCLwY$. We make two observations from the results in Table \ref{tab:sup_mimic-iii}. First, regardless of the label used for training, results were similar on all tasks and competitive with end-to-end. Second, we managed to achieve a stable training even though $\alpha=1.0$ by considering as positive the sample with the same label and close temporally. However, we under-perform compared to $\NCLY$, suggesting that using the $\ltotal$ objective is a better alternative in this case.

\begin{table}[th!]

\caption{Supplementary results on MIMIC-III when using other neighborhood function for supervised approach $\NCLwY$. (D) and (L) stands for Decompensation and Length-of-Stay indicating which labels were used at training. For $\NCLwY$ we used a trade-off parameter $\alpha=1.0$}
\vspace{5pt}
    \centering
\begin{tabular}{l|ll|ll||ll}
\toprule
Task & \multicolumn{4}{|c||}{Decompensation} & \multicolumn{2}{c}{Length-of-stay} \\
\midrule

Metric & \multicolumn{2}{|c}{AUPRC} & \multicolumn{2}{|c||}{AUROC} & \multicolumn{2}{c}{Kappa} \\
\midrule

Head &          Linear &             MLP &          Linear &             MLP &          Linear &             MLP \\
\midrule
\midrule
End-to-End          &  34.3 $\pm$ 1.1 &  34.2 $\pm$ 0.6 &  90.6 $\pm$ 0.3 & 90.6 $\pm$ 0.2 &  43.3 $\pm$ 0.2 &  \textbf{43.4} $\pm$ 0.2 \\\hline
$\NCLwY$ (D) (Ours) &  31.3 $\pm$ 0.7 &  34.4 $\pm$ 0.6 &  89.4 $\pm$ 0.3 &  90.7 $\pm$ 0.1 &  40.8 $\pm$ 0.3 &  43.2 $\pm$ 0.2 \\
$\NCLwY$ (L) (Ours) &  31.2 $\pm$ 0.5 &  34.0 $\pm$ 0.4 &  89.2 $\pm$ 0.2 &  90.6 $\pm$ 0.1 &  40.6 $\pm$ 0.2 &  43.2 $\pm$ 0.2 \\
$\NCLY$ (D) (Ours)   &  \textbf{37.0} $\pm$ 0.6 &  \textbf{37.1} $\pm$ 0.7 &  90.3 $\pm$ 0.2 &  \textbf{90.9} $\pm$ 0.1 &  40.8 $\pm$ 0.3 &  43.3 $\pm$ 0.2 \\
$\NCLY$ (L) (Ours) &  33.5 $\pm$ 1.0 &  36.0 $\pm$ 0.6 &  88.2 $\pm$ 0.5 &  90.5 $\pm$ 0.2 &  \textbf{43.7} $\pm$ 0.2 &  \textbf{43.8} $\pm$ 0.3 \\

\bottomrule
\end{tabular}
\label{tab:sup_mimic-iii}
\end{table}

\subsection{Fine-tuning representation}

 The preferred approach to compare representations is to use a frozen classifier. Contrary to fine-tuning, it preserves what was learned at the training step, allowing a fair comparison. However, if one is interested in the absolute performance on a downstream task, fine-tuning the representation encoder with the classification head usually yields better results. We show in Table \ref{tab:sup_ft_mimic3} that for MIMIC-III, we observe this effect by further improving our unsupervised method. However, as shown in Table \ref{tab:sup_ft_physionet2019}, on Physionet 2019 where our unsupervised method already improved over end-to-end training, performances are degraded. Moreover, we observe that the existing gap between classification heads disappears when fine-tuning. It highlights that fine-tuning shouldn't be used to compare representations learning approaches.

\begin{table}[th!]
\caption{Fine-tuning results for MIMIC-III. The results are averaged over the same 20 runs as frozen evaluation.} 
\vspace{5pt}
    \centering
\begin{tabular}{l|ll|ll||ll}
\toprule
Task & \multicolumn{4}{|c||}{Decompensation} & \multicolumn{2}{c}{Length-of-stay} \\
\midrule

Metric & \multicolumn{2}{|c}{AUPRC} & \multicolumn{2}{|c||}{AUROC} & \multicolumn{2}{c}{Kappa} \\
\midrule

Head &          Linear &             MLP &          Linear &             MLP &          Linear &             MLP \\
\midrule
\midrule
End-to-End            &  34.3 $\pm$ 1.1 &  34.2 $\pm$ 0.6 &  90.6 $\pm$ 0.3 &  90.6 $\pm$ 0.2 &  43.3 $\pm$ 0.2 &  43.4 $\pm$ 0.2 \\
NCL$(n_w)$ (Ours)     &  36.7 $\pm$ 0.5 &  36.6 $\pm$ 0.4 &  91.1 $\pm$ 0.1 &  91.3 $\pm$ 0.2 &  43.7 $\pm$ 0.3 &  43.9 $\pm$ 0.3 \\
NCL$(n_y)$ (D) (Ours) &  36.7 $\pm$ 0.7 &  37.1 $\pm$ 0.7 &  90.7 $\pm$ 0.2 &  90.9 $\pm$ 0.1 &  44.0 $\pm$ 0.2 &  44.0 $\pm$ 0.3 \\
NCL$(n_y)$ (L) (Ours) &  34.8 $\pm$ 1.1 &  34.7 $\pm$ 1.0 &  90.2 $\pm$ 0.2 &  90.3 $\pm$ 0.3 &  42.5 $\pm$ 0.3 &  42.8 $\pm$ 0.3 \\
\bottomrule
\end{tabular}
\label{tab:sup_ft_mimic3}
\end{table}

\clearpage

\begin{table}[h]
\caption{Fine-tuning results for Physionet 2019. The results are averaged over the same 20 runs as frozen evaluation.} 
\vspace{5pt}
    \centering
\begin{tabular}{l|ll|ll||ll}
\toprule
Task & \multicolumn{6}{|c|}{Sepsis} \\
\midrule

Metric & \multicolumn{2}{|c}{AUPRC} & \multicolumn{2}{|c||}{AUROC} & \multicolumn{2}{c}{Utility} \\
\midrule

Head &          Linear &             MLP &          Linear &             MLP &          Linear &             MLP \\
\midrule
\midrule
End-to-End        &  7.6 $\pm$ 0.2 &  8.1 $\pm$ 0.4 &  78.9 $\pm$ 0.3 &  78.8 $\pm$ 0.4 &  27.9 $\pm$ 0.8 &  27.5 $\pm$ 1.0 \\
NCL$(n_w)$ (Ours) &  8.8 $\pm$ 0.4 &  8.9 $\pm$ 0.4 &  80.6 $\pm$ 0.4 &  80.7 $\pm$ 0.3 &  30.2 $\pm$ 0.9 &  30.3 $\pm$ 0.9 \\
NCL$(n_y)$ (Ours) &  8.9 $\pm$ 0.4 &  9.5 $\pm$ 0.4 &  80.6 $\pm$ 0.3 &  80.9 $\pm$ 0.2 &  30.5 $\pm$ 0.7 &  31.6 $\pm$ 0.7 \\
\bottomrule
\end{tabular}
\label{tab:sup_ft_physionet2019}
\end{table}

\subsection{Visualizing Aggregation Impact}
\begin{figure}[h!]
    \centering
    \includegraphics[width=\columnwidth]{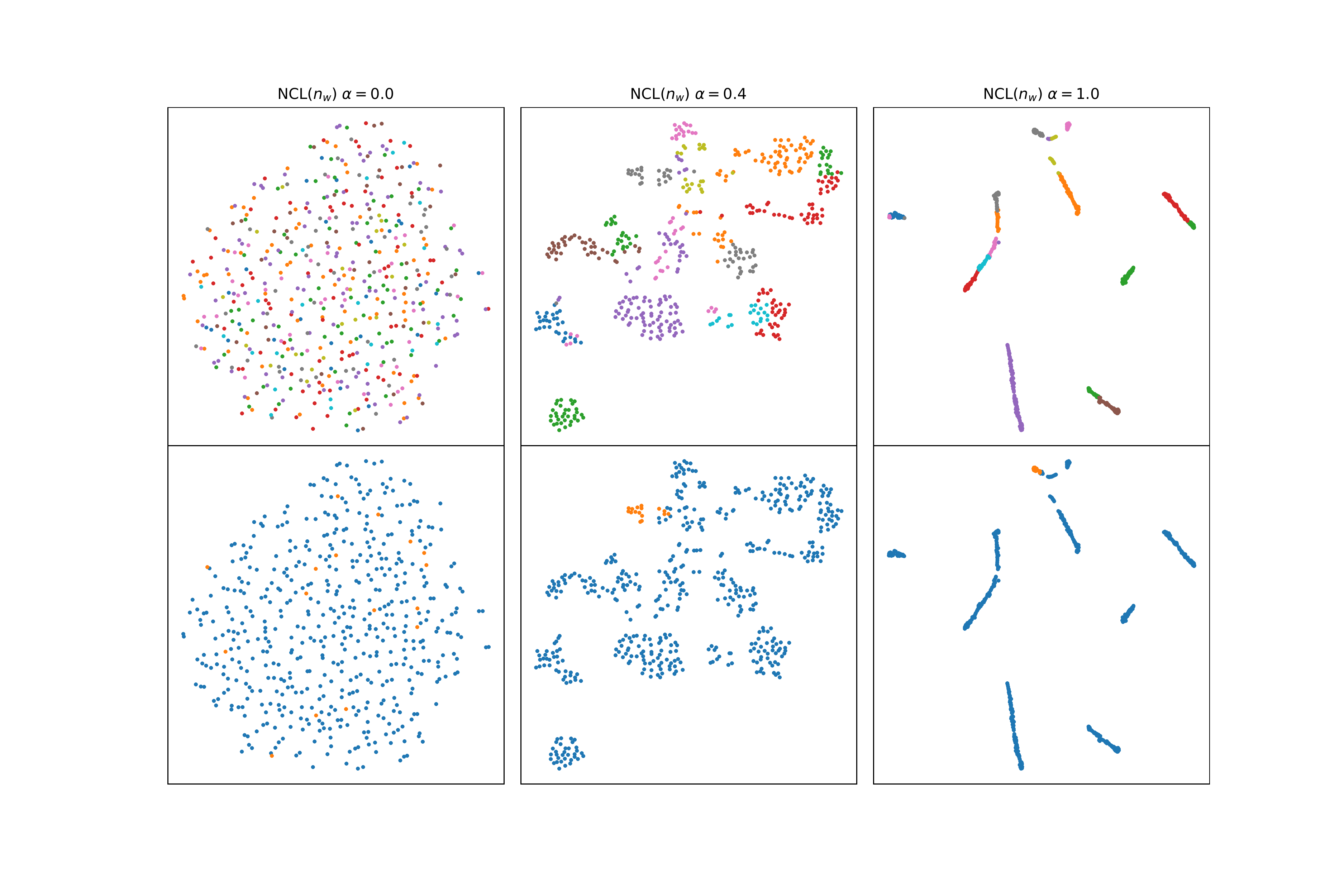}
    \caption{T-SNE plot \citep{mikolov2013efficient} of learned representations on MIMIC-III Benchmark dataset for different values of $\alpha$. (Top row) Labeled per patient. (Bottom row) Labeled with \textit{Decompensation} task. We observe that as conjectured a trade-off in neighbors aggregation is obtained where taking an intermediate value for $\alpha$. }
    \label{fig:tsne}
\end{figure}

In Figure \ref{fig:tsne}, using T-SNE plots we highlight that by increasing $\alpha$ in $\ltotal$, we gradually increase aggregation among neighbors. As conjectured, using only $\lagg$ or $\ldisc$. yields either a collapsed representation or a patient-independent representation.
\onecolumn

\end{document}